\documentclass[twoside,11pt]{article}

\usepackage{blindtext}

 \usepackage[preprint]{jmlr2e}

\usepackage{float}
\usepackage{graphicx}
\usepackage{bm}
\usepackage{multirow}
\usepackage{wrapfig}
\usepackage{subcaption}
\usepackage{mathtools}
\usepackage{graphicx}
\usepackage{caption}
\usepackage{tikz}
\usetikzlibrary{shapes.geometric}
\usetikzlibrary{decorations.pathreplacing,calligraphy}
\usetikzlibrary {shapes.symbols} 
\usetikzlibrary{plotmarks}

\usepackage[utf8]{inputenc} 
\usepackage[T1]{fontenc}    
\usepackage{url}            
\usepackage{booktabs}       
\usepackage{amsfonts}       
\usepackage{nicefrac}       
\usepackage{microtype}      
\usepackage{xcolor}         
\usepackage[skins]{tcolorbox}

\usepackage{lastpage}
\begin{document}








\ShortHeadings{Neural Additive Image Model: Interpretation through Interpolation}{Reuter and Thielmann}
\firstpageno{1}

\title{Neural Additive Image Model:\\Interpretation through Interpolation}

\author{\name Arik Reuter \email arik.reuter@campus.lmu.de \\
       \addr Data Science Working Group \\
       Institute of Mathematics \\
       Clausthal University of Technology\\
       8678 Clausthal-Zellerfeld, Germany
       \AND
       \name Anton Thielmann \email anton.thielmann@tu-clausthal.de \\
       \addr Data Science Working Group \\
       Institute of Mathematics \\
       Clausthal University of Technology\\
       8678 Clausthal-Zellerfeld, Germany
       \AND
       \name Benjamin Saefken \email benjamin.saefken@tu-clausthal.de\\
       \addr Data Science Working Group \\
       Institute of Mathematics \\
       Clausthal University of Technology\\
       8678 Clausthal-Zellerfeld, Germany}

\maketitle

\begin{abstract}
Understanding how images influence the world, interpreting which effects their semantics have on various quantities and exploring the reasons behind changes in image-based predictions are highly difficult yet extremely interesting problems. 
By adopting a holistic modeling approach utilizing Neural Additive Models in combination with Diffusion Autoencoders, we can effectively identify the latent hidden semantics of image effects and achieve full intelligibility of additional tabular effects.
Our approach offers a high degree of flexibility, empowering us to comprehensively explore the impact of various image characteristics. We demonstrate that the proposed method can precisely identify complex image effects in an ablation study.  To further showcase the practical applicability of our proposed model, we conduct a case study in which we investigate how the distinctive features and attributes captured within host images exert influence on the pricing of Airbnb rentals.
\end{abstract}


\section{Introduction}
Over the last decade, Deep Neural Networks (DNN) have achieved remarkable results in a variety of tasks. Text classification \citep{kowsari2019text}, Speech Recognition (e.g. \citep{radford2022robust, ma2022visual}), Computer Vision \citep{CV2020deep}, Genome Detection \citep{danaee2017deep} or Protein Folding \citep{jumper2021highly} are just a few areas in which DNNs have considerably exceeded their predecessors, sometimes by astonishingly wide margins. 
The flexibility and predictive power of DNNs, however, often comes at the price of inherently uninterpretable model predictions, leading to DNNs not being used in high risk domains.

Achieving a comprehensive understanding of the inner workings and decision-making processes of these complex models has significant implications for their practical deployment. Various techniques have been proposed to enhance the interpretability of neural networks, including post-hoc methods, as well as a-priori approaches. 
To differentiate between different forms of model interpretability, we follow the literature (e.g. \citep{zhang2021survey, fan2021interpretability}) and distinguish between Local and Global interpretability as well as Post-hoc and A-priori interpretability. 
Local Interpretable Model-Agnostic Explanations (LIME) \citep{ribeiro2016should}, Shapley values \citep{shapley1953quota} or Layer-wise Relevance Propagation (LRP) \citep{bach2015pixel} for example, are post-hoc and locally interpretable methods that are especially well-suited for identifying relevant pixel areas in computer vision related tasks. A-priori and global explanations of model predictions are less commonly adapted and often come with certain disadvantages \citep{zhang2021survey}. For instance, \citet{zhang2018interpretable} introduced an a-priori interpretable Convolutional Neural Network (CNN), which is only applicable to classification tasks. 
For tabular data, Neural Additive Models (NAMs) \citep{agarwal2021neural} are a-priori globally interpretable models that are closely related to Generalized Additive Models (GAMs) \citep{hastie2017generalized, wood2017generalized}. NAMs allow to visualize individual feature effects and can incorporate higher order feature interactions. For example, \citet{xu2023coxnam} in survival analysis and \citet{jo2022neural} in time-series prediction have demonstrated the practical and theoretical relevance of NAMs. \citet{rugamer2022semi} extend NAMs to multi-modal data and incorporate images into their modeling framework. 
However, the unstructured components still lack interpretability, leaving them difficult to comprehend. Therefore, while there exist various model structures that can provide a-priori global interpretability, they are confined to a specific singular data type.

In the domain of computer vision, the recent advancements in generative modelling based on diffusion models allow to produce highly realistic and detailed images across a wide range of styles and subjects \citep{sohl2015deep, ho2020denoising}. These models operate by gradually constructing an image through a process that starts with random noise and progressively refines it into a coherent output, simulating the reverse of a diffusion process. This approach has demonstrated remarkable capabilities in generating images that are not only visually compelling but also exhibit a high degree of diversity and control over the generation process \citep{dhariwal2021diffusion, nichol2021glide}. 
Applications of diffusion models extend beyond image synthesis to tasks such as image-to-image translation, super-resolution, and inpainting, showcasing their versatility and potential to push the boundaries of what is achievable in generative visual content creation \citep{saharia2022palette, yue2024resshift, lugmayr2022repaint}.

By utilizing Diffusion Autoencoders (DAEs), a variant Diffusion Probabilistic Models that allows to encode and decode images into a semantically meaningful embedding space \citep{preechakul2022diffusion}, we are able to comprehensively interpret image effects in an additive modelling framework, that incorporates numerical, as well as image-covariates. 

Multi-modal data, comprising tabular data as well as images,  is not only common in everyday life, but also prevalent in high-risk domains, such as human-resource-management and healthcare. For instance, combining images and, e.g., patient information in tabular form can facilitate a more comprehensive understanding of medical decision-making processes. In summary, this paper aims to make the following contributions:

\begin{itemize}
    \item We present a model structure that seamlessly incorporates tabular data and images while preserving global interpretability through the utilization of additivity constraints.
    \item Our framework for understanding tabular data and images is highly intuitive, flexible, and provides profound insights into the data.

    \item We experimentally validate the effectivenes of the NAIM framework in identifying the effects of numerical features combined with image-covariates in an ablation study with synthetic data. 
    
    \item We showcase the effectiveness of the approach and present both global feature effects and locally interpretable examples in a case study.
\end{itemize}

\begin{figure}
    \centering
    \includegraphics[width=.95\textwidth]{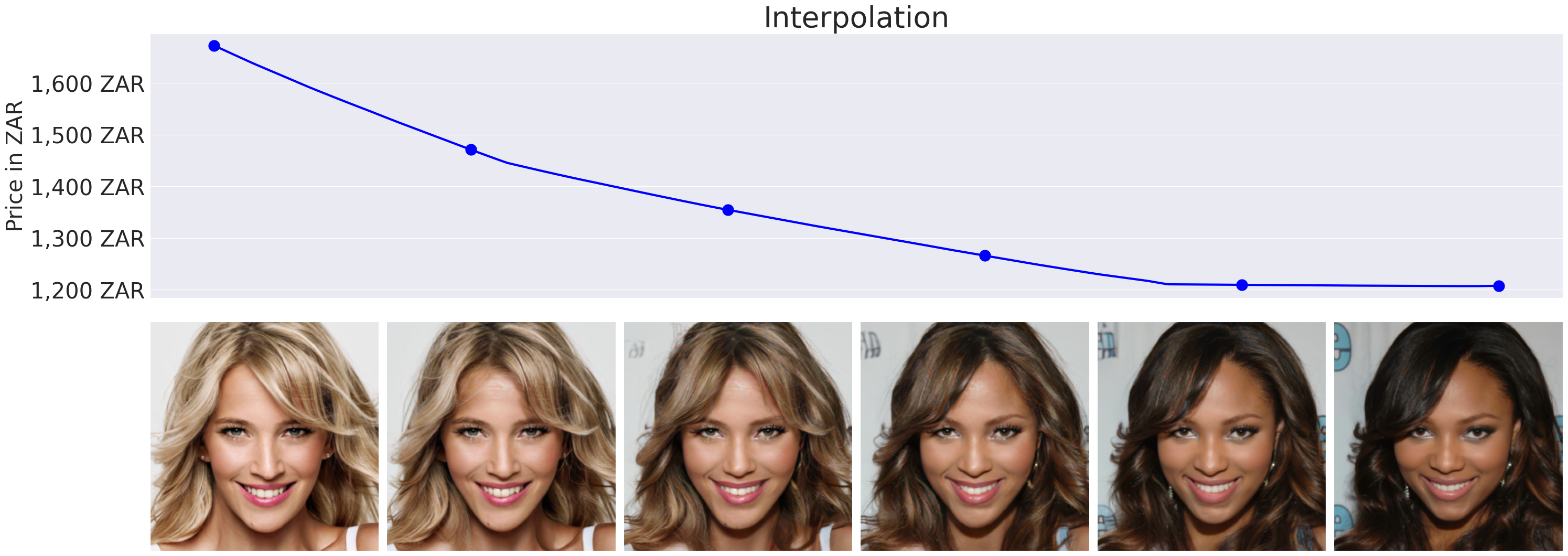}
    \caption{Effect on the price of an Airbnb listing obtained by interpolating between two input images. A Neural Additive Model that includes an image effect is used to predict the expected price. Image Interpolating is achieved by linearly interpolating between image representations in the embedding space of a Diffusion Autoencoder.}
    \label{fig:interpolation_skincolor}
\end{figure}

\section{Neural Additive Image Model}

Let $\mathcal{D} = \{ (\bm{x}^{(i)}, y^{(i)})\}_{i=1}^n$ be a dataset of size n. Let further $y \in \mathbb{R}$ denote the target variable that can be arbitrarily distributed and $J$ denote the number of features such that each input $\bm{x} = (x_1, x_2, \dots, x_J) \in \mathbb{R}^J$ contains $J$ features. Given a 
link function $g : \mathbb{R} \rightarrow \mathbb{R}$, a GAM in its most fundamental form can be expressed as:
\begin{equation}
\label{eq:GAM}
    g\left(\mathbb{E}\left[y|\bm{x}\right]\right) =  \beta_0 + \sum_{j}^{J} f_j(x_j),
\end{equation}
\noindent where $\beta_0$ denotes the global intercept and $f_{j} : \mathbb{R} \rightarrow \mathbb{R}$
denote the univariate shape functions corresponding to input feature $x_j$. In classical GAMs $f_j\in\mathcal{C}^2(\mathbb{R})$, i.e. the shape functions are smooth, two times differentiable functions, e.g., cubic splines and the parameters are optimized via a penalized least squares criterion \citep{hastie2017generalized, wood2017generalized} . The effects of features on the target variable are finally interpreted by, e.g., plotting the graph of the shape function. Pairwise feature interactions are often accounted for in the form of:
\begin{equation}
\label{eq:G^2AM}
    g\left(\mathbb{E}\left[y|\bm{x}\right]\right) =  \beta_0 + \sum_{j}^{J} f_j(x_j) + \sum_{j, k: j\neq k}^{J} f_{jk}(x_j, x_k),
\end{equation}
\noindent where $f_{jk} : \mathbb{R}^2 \rightarrow \mathbb{R}$ denote all feature interactions between input features $(x_j,x_k)\in\mathbb{R}^2$. 


\citet{nori2019interpretml} introduced tree-based, cyclic gradient boosting GAMs, resulting in much more powerful, yet interpretable GAMs. Similarily, \citet{chang2021node} used differentiable decision trees. \citet{dubey2022scalable} introduced scalable polynomials as shape functions.

Neural Additive Models (NAMs) on the other hand leverage multilayer perceptrons (MLPs) as shape functions while preserving the overall additive model structure \citep{agarwal2021neural}.
This enables the development of interpretable and powerful models that leverage the advantages inherent to the deep learning paradigm.
Thus, higher order feature interactions can be accounted for in various ways, by, e.g, adding higher-dimensional input MLPs directly into Equation \ref{eq:G^2AM} \citep{enouen2022sparse, kim2022higher}. 
Through the inherent flexibility of MLPs, further adaptations, e.g. shared learned basis functions \citep{radenovic2022neural, enouen2022sparse} or distributional approaches \citep{thielmann2023neural} are possible. 

Additionally, independent of the shape function, fitting the model with gradient descent allows for other data types  to be incorporated into the modelling, that traditional GAMs are not able to account for. While all GAM adaptations achieve remarkable results in terms of predictive performance, only \citet{rugamer2022semi} incorporated images into their GAM. 
However, the unstructured parts remain completely uninterpretable. 

To account for interpretable image effects, we adapt Equation \ref{eq:G^2AM} accordingly. Let further $\bm{x}$ denote the tabular features, where $x_j$ is the $j$-th tabular feature, and $\bm{x}_{img}$ denotes the images. 
 Equation \ref{eq:G^2AM} can subsequently be adapted to:
\begin{equation}
\label{eq:G^2AM+img}
    g(\mathbb{E}\left[y|\bm{x}, \bm{x}_{img} \right]) =  \beta_0 + \sum_{j}^{J} f_j(x_{j}) + \sum_{j, k: j \neq k}^{J} f_{jk}(x_{j}, x_{k}) + f_{img}(\bm{E}_\phi(\bm{x}_{img})),
\end{equation}
where $\bm{E}_\phi$ denotes a CNN with parameters $\phi$ that encodes an input image into a latent representation $\bm{z} = \bm{E}_\phi(\bm{x}_{img}) \in \mathbb{R}^l$. This latent representation is subsequently processed by the image-specific shape function $f_{img}: \mathbb{R}^l \rightarrow \mathbb{R}$. Note that Equation \ref{eq:G^2AM+img} can be directly adapted to include interactions between numerical and image covariates.

\subsection{Interpretable Image Representations}
\label{sec:DAEs}

Diffusion Autoencoders (DAEs) \citep{preechakul2022diffusion} are a powerful class of autoencoders that can provide semantically meaningful and almost perfectly decodable latent representations of images. 
The key innovation of DAEs is the use of a semantic encoder in combination with a Denoising Diffusion Implicit Model (DDIM) \citep{song2020denoising}. The semantic encoder captures high-level features of an input image while the DDIM acts as both an encoder of more subtle stochastic variations in the input image and as a powerful decoder. 
More specifically, the semantic encoder $\bm{E}_\phi$
provides a semantic subcode $\bm{z}  = \bm{E}_\phi(\bm{x}_{img})$ that encodes primarily the high-level information for a given input image $\bm{x}_{img}$. The DDIM, on the other hand, can map a stochastic subcode $\bm{x}_{img}^{(T)}$, for instance, sampled from a standard-normal distribution, onto a real-world image. 

\begin{figure}[H]
    \centering
    \includegraphics[width=.9\textwidth]{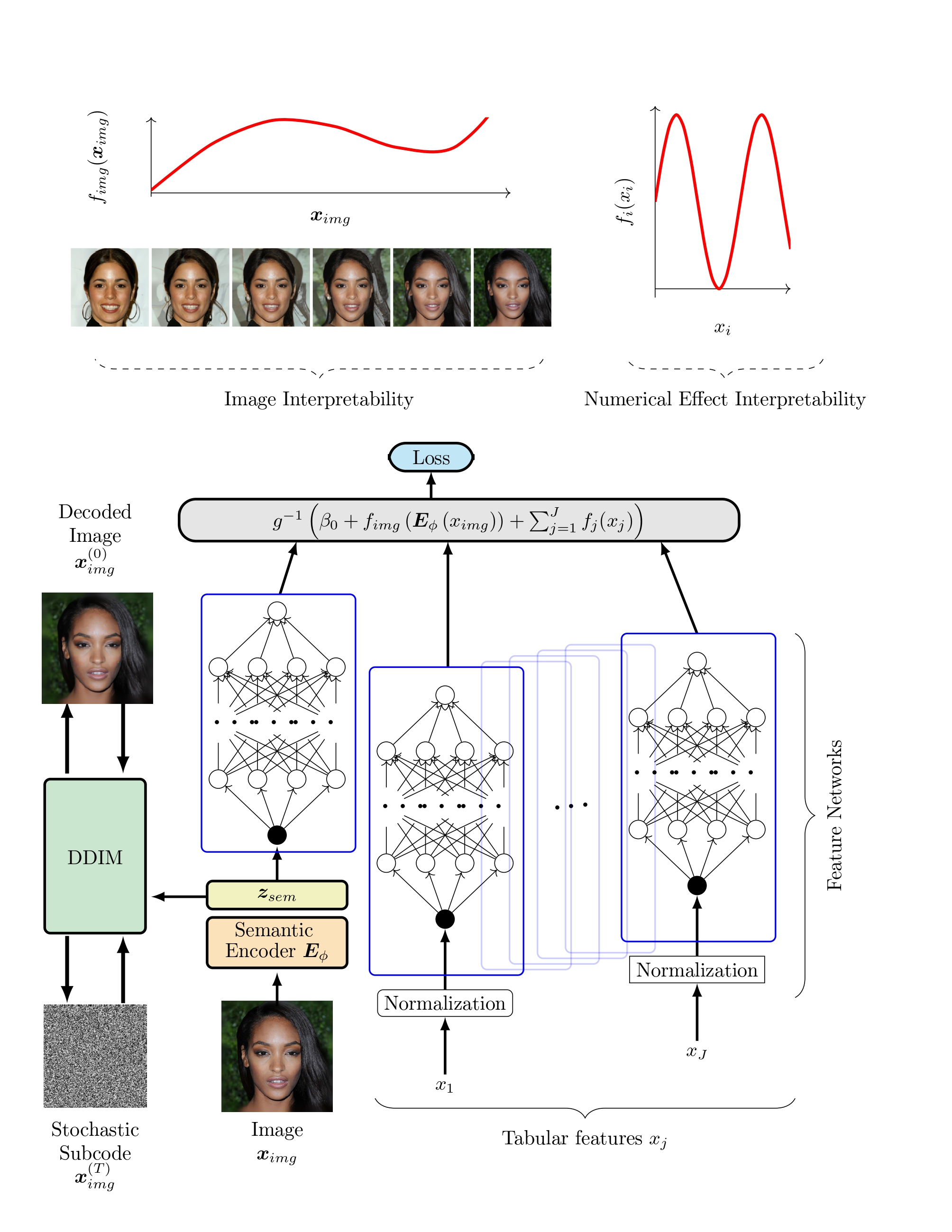}

\caption{Model architecture of the base version of the proposed approach not accounting for interaction effects. The tabular data is fit as independent MLPs, the images are encoded with an CNN encoder and fit with a separate MLP. All feature effects are summed up for the model prediction.}
\end{figure}

This is achieved by denoising the noise-input in a step-wise manner until the final image is generated. The reverse diffusion process that is used to generate the image from the stochastic subcode can be described as follows: 
\begin{equation}
\label{eq:ReverseDiffusion}
    p_{\psi}(\bm{x}_{img}^{(0:T)}| \bm{z}) = p(\bm{x}_{img}^{(T)}) \prod_{t=1}^T p_{\psi}(\bm{x}_{img}^{(t-1)}|\bm{x}_{img}^{(t)}, \bm{z}),
    \end{equation}

\noindent where $\psi$ denotes the parameters of the DDIM and the semantic subcode $\bm{z}$ is used to condition the image generation on the high-level semantics of the reconstructed image.
Further,  $p_{\psi}(\bm{x}_{img}^{(t-1)}|\bm{x}_{img}^{(t)}, \bm{z})$ is given by:
$$
\label{eq:step2}
    p_{\psi}(\bm{x}_{img}^{(t-1)}|\bm{x}_{img}^{(t)}, \bm{z}) = \begin{cases} 
\mathcal{N}(f_{\psi}(\bm{x}^{(1)}, 1, \bm{z}), 0) & \text{if } t = 1 \\
q(x_{img}^{(t-1)} | \bm{x}_{img}^{(t)}, f_{\psi}(\bm{x}_{img}^{(t)}, t, \bm{z})) & \text{otherwise}.
\end{cases}
$$
$f_{\psi}$ is parameterized as a noise prediction network as done in \citet{preechakul2022diffusion} and $q(\bm{x}_{img}^{(t-1)}|\bm{x}_{img}^{(t)}, \bm{x}_{img}^{(0)})$ defines the inference distribution matched by the reverse generative process (\ref{eq:ReverseDiffusion}).

Since DDIMs have a deterministic generative process, they can be interpreted as autoencoders, which can not only decode the stochastic code $\bm{x}_T$ into an image, but also encode an image into a corresponding stochastic (sub-)code by running the generative process backwards. For diffusion autoencoders, the encoding procedure is enhanced by also conditioning on $\bm{z}$, which does not only improve image generation by providing the compact semantic information in $\bm{z}$, but thereby also increases inference speed. 

The embedding space of the semantic subcode has several advantageous properties: 
First, every semantic subcode can be decoded and thus visualized.
Second, simple linear interpolation between semantic subcodes corresponds to meaningful semantic interpolation in the space of real images. Third, one can obtain interpretations of some directions in the embedding space. Those directions can be, for instance, given by the normal vector of the separating hyperplane corresponding to a linear classifier trained on the semantic codes. 

\subsection{Image-Effect Interpretation through Interpolation}
\label{section:Image_effect_interpretation}

\begin{figure}[h]
   \centering
   \includegraphics[width=.9\textwidth]{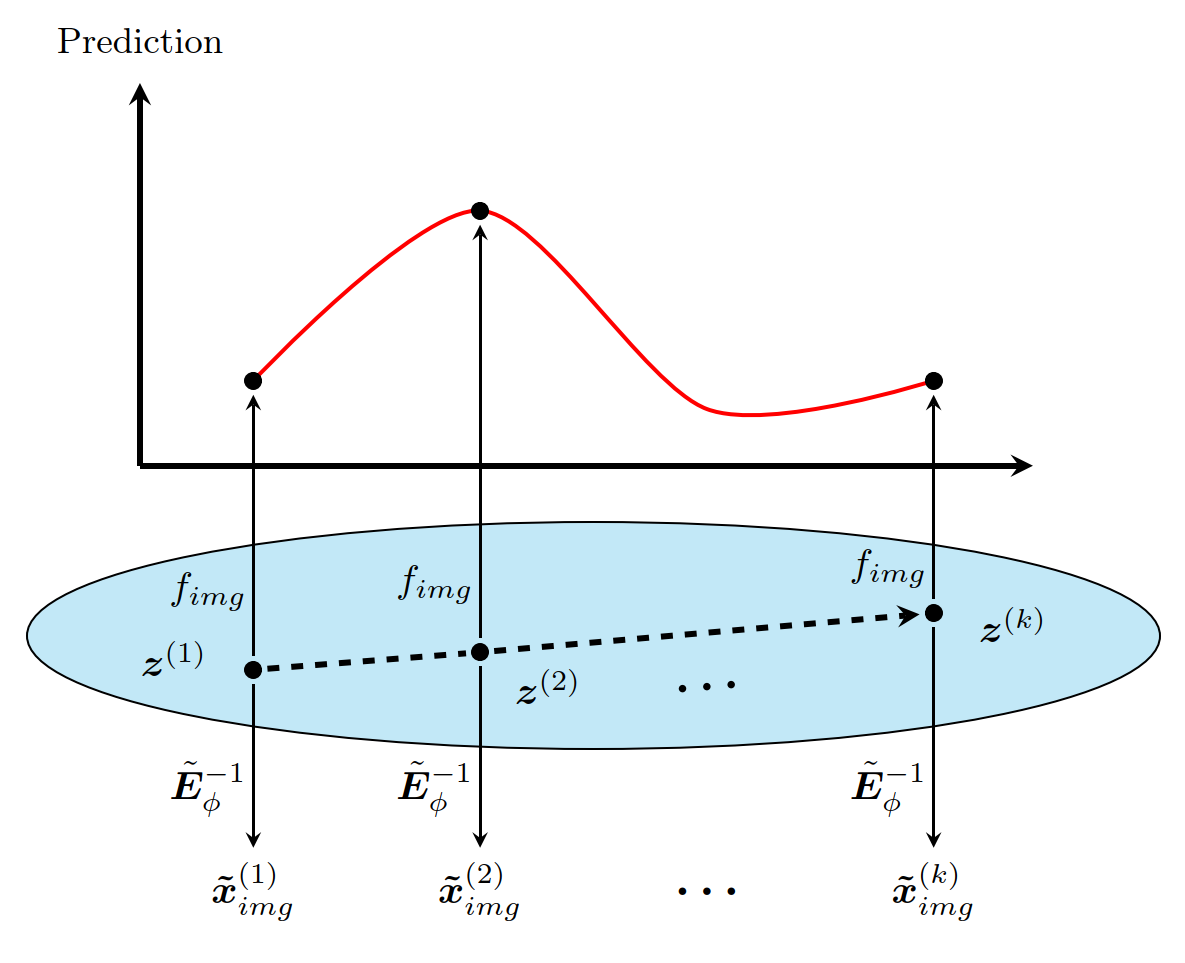}

\caption{Interpolating between the latent semantic codes $\bm{z}_{int}^{(1)}$ and $\bm{z}_{int}^{(k)}$ in the embedding space can be visualize in pixel-space as the sequence $\left( \bm{\tilde{x}}_{img}^{(1)},\bm{\tilde{x}}_{img}^{(2)}, \ldots, \bm{\tilde{x}}_{img}^{(k)} \right)$ by using the decoder $\tilde{\bm{E}}_{\phi}^{-1}$ on the latent codes. Each latent code $\bm{z}_{int}$, on the other hand, also allows to make the prediction $f_{img}(\bm{z}_{int})$ for a response variable of interest. Plotting the predictions $\left( f_{img}(\bm{z}_{int}^{(1)}), f_{img}(\bm{z}_{int}^{(2)}), \ldots, f_{img}(\bm{z}_{int}^{(k)}) \right)$ against the generated images $\left( \bm{\tilde{x}}_{img}^{(1)},\bm{\tilde{x}}_{img}^{(2)}, \ldots, \bm{\tilde{x}}_{img}^{(k)} \right)$  allows to investigate the image effect.}
\label{fig:enter-label} 
\end{figure}

NAMs provide a framework to globally interpret the effects of continuous features by plotting the graph of the shape functions $f_j$ corresponding to the respective univariate (or sometimes bivariate) covariates. However, since image-data lacks any inherently semantically interpretable dimensions in pixel-space, one has to rely on meaningful image representations, their semantics and interpretations, in order to achieve similar understanding of image data. An important, and often existing property, of those learned representations is \textbf{semantic continuity}, i.e. representations that are close in their embedding space are also similar in terms of their semantics, as understood by humans \citep{oh2016deep, NIPS2016_90e13578, bell2015learning}. One can formalize this by assuming a bijective function $m: \mathbb{R}^p \rightarrow \mathbb{R}^l$ that maps a ground-truth semantic feature $\Phi(\bm{x}_{img})$ of an image $\bm{x}_{img}$ onto the embedding $\bm{E}_\phi(\bm{x}_{img})$ of that image, i.e. $m = \bm{E}_\phi \circ \Phi^{-1} $ if $ \Phi$ is invertable. Examples for those ground-truth features obtained by $\Phi$ could be the average red-value of all pixels in an RGB-image, the age of a portrayed person or the position of an object. We can now define semantic continuity as assuming regularity, i.e., continuity in the mathematical sense, or more general different degrees of smoothness for $m$. 

To obtain human insight into the mostly incomprehensible nature of representations obtained by deep learning methods, a straightforward requirement is \textbf{decodability}. This property of embeddings implies that there is not only a way to encode an image into a semantic representation, via $\bm{z} = {\bm{E}}_\phi(\bm{x}_{img})$ but also always the possibility to translate any latent code back to the pixel level, i.e. an image which can be intuitively understood by humans. This can be interpreted as being able to approximate the inverse of the encoding function ${\bm{E}}_\phi$ via ${\tilde{\bm{E}}}^{-1}_\phi \approx {\bm{E}}^{-1}_\phi$. One can note that in the context of DAEs \citep{preechakul2022diffusion}, embeddings do not only posses great properties in terms of semantic continuity, but are seminal in terms of near-perfect decodability.

For the purpose of image-effect interpretation, semantic continuity allows to locally change embeddings in a semantically meaningful way, while the decodability ensures that those changes can be visualized. 
A sequence of latent representation $\left(\bm{z}^{(i)} \right)_{i=1}^k \vcentcolon = \left( \bm{z}^{(1)}, \bm{z}^{(2)}, \ldots, \bm{z}^{(k)} \right)$ then corresponds to a series of generated images. If the augmented latent representations are also used to make predictions, one can visualize the effect of changing the embeddings by plotting the prediction against the augmented images.
In the NAIM framework, this means that we can plot the effect $f_{img}(\bm{z}^{(i)})$ against images ${\tilde{\bm{E}}}^{-1}_\phi(\bm{z}^{(i)})$ for a selection of $k$ shifted embeddings $\left(\bm{z}^{(i)} \right)_{i=1}^k$. More precisely, the idea is to plot tuples of images, based on decoded embeddings, and corresponding predictions $\left({\tilde{\bm{E}}}^{-1}_\phi(\bm{z}^{(i)}), f_{img}(\bm{z}^{(i)}) \right)_{i=1}^k$ based on a sequence  $\left(\bm{z}^{(i)} \right)_{i=1}^k$. Thus, we can visualize a (semi-)continuous feature effect of image semantics provided a sequence of latent variables.

The following result shows that interpolation, and in particular linear interpolation, is a reasonable approach for obtaining such a sequence of latent representations $\left(\bm{z}^{(i)} \right)_{i=1}^k$.

{\bf Theorem 1}{
\label{theorem1}
For $h : \mathbb{R}^l \rightarrow \mathbb{R}^p$ a differentiable function one has that \begin{equation}
h \left(\lambda \bm{z} + (1-\lambda) \tilde{\bm{z}} \right) = \lambda h(\bm{z}) + (1- \lambda) h(\tilde{\bm{z}}) + (1 - \lambda)R
\end{equation}
for any $\bm{z}, \tilde{\bm{z}} \in \mathbb{R}^l$ with $0 \leq \lambda \leq 1$ and $R = \bm{o}(\lvert \lvert \bm{z} - \tilde{\bm{z}}  \lvert \lvert)$. A proof for this theorem can be found in appendix \ref{proof_theorem}.
}
Note that the approximation becomes (locally) exact if $h$ is (locally) affine. 
This theorem demonstrates that first forming the convex combination between two sufficiently close embeddings $\bm{z} = \bm{E}_\phi(\bm{x}_{img})$ and $\tilde{\bm{z}} = \bm{E}_\phi(\Tilde{\bm{x}}_{img})$ followed by computing ground-truth semantic features $\Phi(\bm{x}_{img}) = m^{-1}(\bm{z})$ and $\Phi(\Tilde{\bm{x}}_{img}) = m^{-1}(\tilde{\bm{z}})$ is approximately equal to first computing the semantic features and then forming the convex combination between the semantic features $m^{-1}(\bm{z})$ and $m^{-1}(\tilde{\bm{z}})$. (Note that $m^{-1} = \Phi \circ \bm{E}^{-1}_\phi$). More specifically for the NAIM framework, if one has learned an invertable and smooth function $\bm{E}_\phi$, that respects semantic features in $\mathbb{R}^p$ well, we can expect that interpolating between two sufficiently close embeddings $\bm{z}$ and $\tilde{\bm{z}}$ followed by decoding with $\tilde{\bm{E}}^{-1}_\phi$ yields the same kind of interpolation for the ground-truth semantic features:
\begin{equation}
\Phi(\tilde{\bm{E}}^{-1}_\phi(\left(\lambda \bm{z} + (1-\lambda) \tilde{\bm{z}} \right)) \approx \lambda \Phi(\tilde{\bm{E}}^{-1}_\phi(\bm{z})) + (1- \lambda)\Phi(\tilde{\bm{E}}^{-1}_\phi(\tilde{\bm{z}})).
\end{equation}

We propose two ways of augmenting images for the purpose of obtaining latent sequences $\left(\bm{z}^{(i)} \right)_{i=1}^k$ with the ultimate goal of image effect interpretation:
First, to use image interpolation to locally understand how changes in the image influence the prediction for the response variable. This has the benefit that the effect of arbitrary attributes can be analyzed by selecting pairs of examples where one image has a certain attribute and the other does not. In the context of DAEs, this can be achieved by selecting two images, $\bm{x}_{img}$ and $\bm{x'}_{img}$, computing $\bm{z} = \bm{E}_\phi(\bm{x}_{img})$ and $\bm{z '} = \bm{E}_\phi(\bm{x'}_{img})$ followed by obtaining the latent sequence $\left(\bm{z}_{int}^{(i)}\right)_{i=1}^k$ as a result of simple linear interpolation between the embeddings $\bm{z}$ and $\bm{z '}$. This means that 
\begin{equation}
    \left(\bm{z}_{int}^{(i)} \right)_{i=1}^k = \left ( \left( 1 - \frac{i-1}{k-1} \right) \bm{z} + \frac{i-1}{k-1} \bm{z'} \right )_{i=1}^k
\end{equation}


\begin{figure}[H]
    \centering
    \includegraphics[width=.95\textwidth]{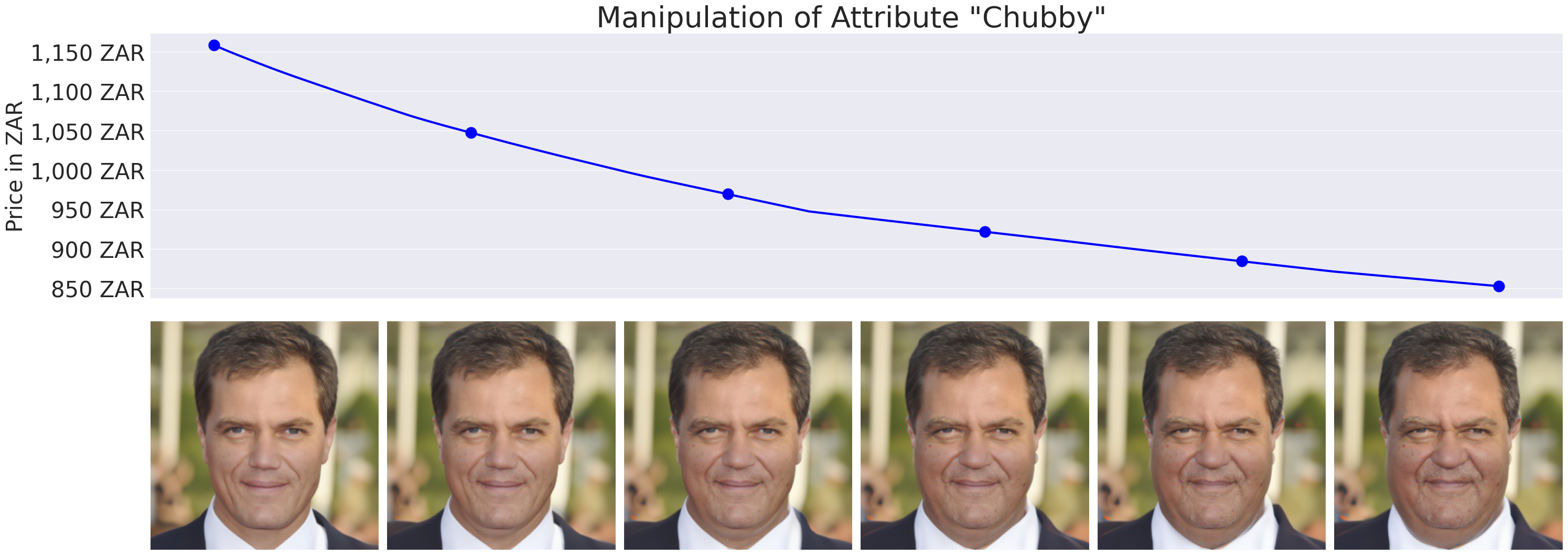}
    \caption{Effect of attribute Manipulation for the feature \textit{Chubby} on the expected rental price. Through (semi) continuous interpolation, an effect trend can be visualized.}
    \label{fig:chubby_plus}
\end{figure}
Second, attribute manipulation, i.e. the explicit change of a specific semantic quality of an image, can be used to investigate the effect of pre-defined attributes. By explicitly changing an attribute of an image, one can plot the local effect of this change, see, e.g., Figure \ref{fig:chubby_plus}. One can also visualize the global effect that manipulating all training samples has on the distribution of the predictions.
Image manipulation can be achieved by shifting the embedding of an input image $\bm{z} = \bm{E}_\phi(\bm{x}_{img})$ into the direction that corresponds to a specific semantic attribute of an image. Given a vector $\bm{v}_{attr}$ that encodes a particular semantic aspect in the embedding space and a maximum shift magnitude $\alpha$, one can construct a sequence of latent codes via manipulation as
\begin{equation}
    \left(\bm{z}_{man}^{(i)} \right)_{i=1}^k = \left ( \bm{z} + \alpha \left(\frac{i-1}{k-1}\right) \frac{\lvert \lvert \bm{z}_{man} \lvert \lvert}{\lvert \lvert \bm{v}_{attr} \lvert \lvert}  \bm{v}_{attr} \right )_{i=1}^k,
\end{equation}
where the ratio of norms ensures the strength of manipulation is roughly similar to $\lvert \lvert \bm{z} \lvert \lvert$. Analogously to the interpolation case, one can now visualize image effects via plotting tuples $\left({\tilde{\bm{E}}}^{-1}_\phi(\bm{z}_{man}^{(i)}), f_{img}(\bm{z}_{man}^{(i)}) \right)_{i=1}^k$.
While a default choice of $\alpha = 1$ seems to work reasonable well, we find that visually inspecting the manipulated images and choosing $\alpha$ accordingly is recommended in practice.
A manipulation vector $\bm{v}_{attr}$ can be obtained, for instance, as the normal vector of a hyperplane that separates embeddings of images of two different classes \citep{preechakul2022diffusion}. 

Note that attribute manipulation can be seen as a special case of interpolation where one interpolates between $\bm{z}_{man}$ and $\bm{z}_{man} + \alpha \frac{\lvert \lvert \bm{z}_{man} \lvert \lvert}{\lvert \lvert \bm{v}_{attr} \lvert \lvert}  \bm{v}_{attr}$.

\section{Experiments with Synthetic Data}

To evaluate the capability of our suggested approach in discovering and disentangling the impacts of multiple metric covariates as well as an additional image covariate, we execute an ablation study utilizing synthetic data. 

\subsection{Data Generation}

\newcommand{\Ds}{\mathcal{D}_{\text{squares}}}
\newcommand{\Dc}{\mathcal{D}_{\text{colors}}}

We generate two synthetic datasets that are generated with several numerical effects as well as an image effect.
The first dataset $\Ds$ ("squares-data") contains images that consist of a white square on grey background such that the white square has half the side length of the total image. The coordinates of the center of the white square are drawn uniformly at random. 
The image-part of the second synthetic dataset $\Dc$ ("colors-data") comprises monochromatic images where the red-, green- and blue-values are drawn uniformly at random.  Let further $\bm{\Phi}_{xval}$ 
denote the simple feature-extraction function that maps an image from $\Ds$ onto the x-coordinate of the center of its white square and $\bm{\Phi}_{red}$ the function that extracts the (average) red value from an image. Note, that we normalize the output of those functions such that $\bm{\Phi}_{xval}$ and $\bm{\Phi}_{red}$ both have the unit-interval $[0,1]$ as their co-domain.
The final image effects are then given by $f_{img} \circ \bm{\Phi}_{xval}$ for $\Ds$ and $f_{img} \circ \bm{\Phi}_{red}$ for $\Dc$ with $f_{img} \in \{f_{{img}_1}, f_{{img}_2}, f_{{img}_3} \}$ where $f_{{img}_1}(x) = 2x$, $f_{{img}_2}(x) = 2x^4$ and $f_{{img}_3}(x) = \sin(2 \pi x)$ (see figure \ref{fig:ablation_tab_effects}).

Besides the image effects, we simulate the effects of the numerical data for $\Ds$ and $\Dc$, by using a linear function  $f_1(x) = 2 x$, a power-function $f_2(x) = x^2$ and a sinusoidal function $f_3(x) = \sin(2 \pi x)$. We apply those effects to x-values drawn uniformly at random from the unit interval. Note, that we choose the same form of effects for the numerical- and image-covariate to not allow the model to identify effects over their mere form. We choose $f_2$ to slightly deviate from $f_{{img}_2}$ for visualization purposes. 
To form the response values, we add up all the effects and add zero-centered normally distributed noise $\epsilon \sim \mathcal{N}(0, 0.1)$.

A dataset size of 100,000 samples is used for training the DAE and the NAIM, while additional 10,000 samples are used for testing. 

For the visualizations, we display images that correspond to decoded semantic codes for 10 interpolation steps. We then plot the true and the learned effect for those generated images. 
To have a reference for the generated images, we also display samples that result from a trivial semantic interpolation based on the features extracted by $\bm{\Phi}_{xval}$ and $\bm{\Phi}_{red}$. To be more precise, for $\Ds$, we simply linearly interpolate the x-value of the center of the white images and generate the corresponding images. For $\Dc$ the reference interpolation is obtained by linearly interpolating the red value between the red value of the start-image and the red value of the end-image.

\subsection{Results}

\begin{table}[t]
\centering
\begin{tabular}{|c|c|c|c|c|}
\hline
\multicolumn{5}{|c|}{Squares Data} \\
\hline
$f_{img}(x)$         & $MSE$ w/o image             & $R^2$ w/o image     & $MSE$ w/ image               & $R^2$ w/ image \\ \hline
$2x$                 & $1.08$                    & $0.380$               & $0.0106$                       & $0.996$ \\
$2x^4$               & $0.953$                   & $0.471$               & $0.0106$                       & $0.996$ \\
$\sin( 2 \pi x)$     & $1.16$                    & $0.405$               & $0.0106$                       & $0.996$ \\ \hline
\end{tabular}

\begin{tabular}{|c|c|c|c|c|}
\hline
\multicolumn{5}{|c|}{Colors Data} \\
\hline
$f_{img}(x)$         & $MSE$ w/o image            & $R^2$ w/o image     & $MSE$ w/ image              & $R^2$ w/ image \\ \hline
$2x$                 & $1.08$                   & $0.383$               & $0.0106$                    & $0.996$ \\
$2x^4$               & $1.03$                   & $0.377$               & $0.0106$                    & $0.996$ \\
$\sin( 2 \pi x)$     & $1.21$                   & $0.357$               & $0.0122$                    & $0.996$ \\ \hline

\end{tabular}
\caption{Overall fit of the NAIM evaluated with a test set of size 10,000 containing image and numerical covariates.}
\label{table:overall_fit}
\end{table}

First, the entire NAIM model is able to almost perfectly predict the correct response-value for the squares data and the colors data since the Mean-Squared-Error of the model is very close to the used noise variance of $0.01$ (Table \ref{table:overall_fit}). 
One can also see that including the image covariate, as opposed to using just a NAM, noticeably improves the predictive capabilities even if the image-effect and the numerical effects are very similar or even identical in their structure and magnitude.

\subsubsection{Numerical Effects}

\begin{figure}[h]
    
  \begin{subfigure}[b]{0.5\linewidth}
    \centering
    \includegraphics[width=0.85\linewidth]{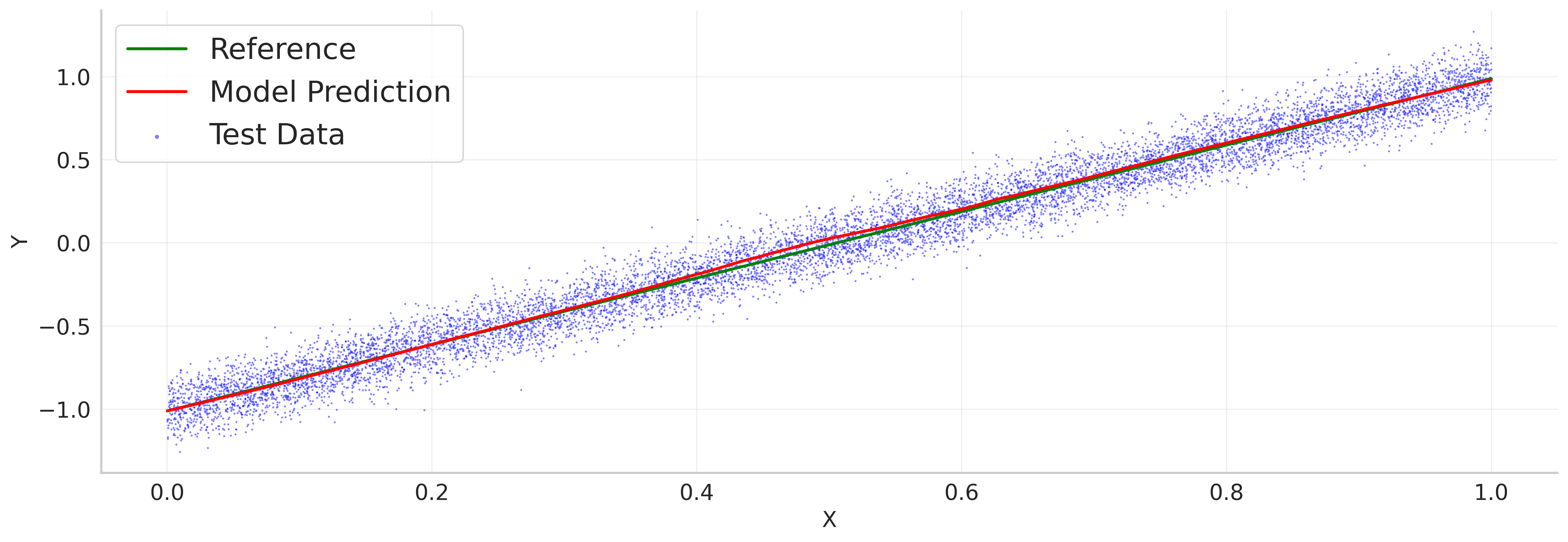}  
    \caption{Effect 1} 
    \label{fig:ablation_tab_effects_1} 
  \end{subfigure}
  \begin{subfigure}[b]{0.5\linewidth}
    \centering
    \includegraphics[width=0.85\linewidth]{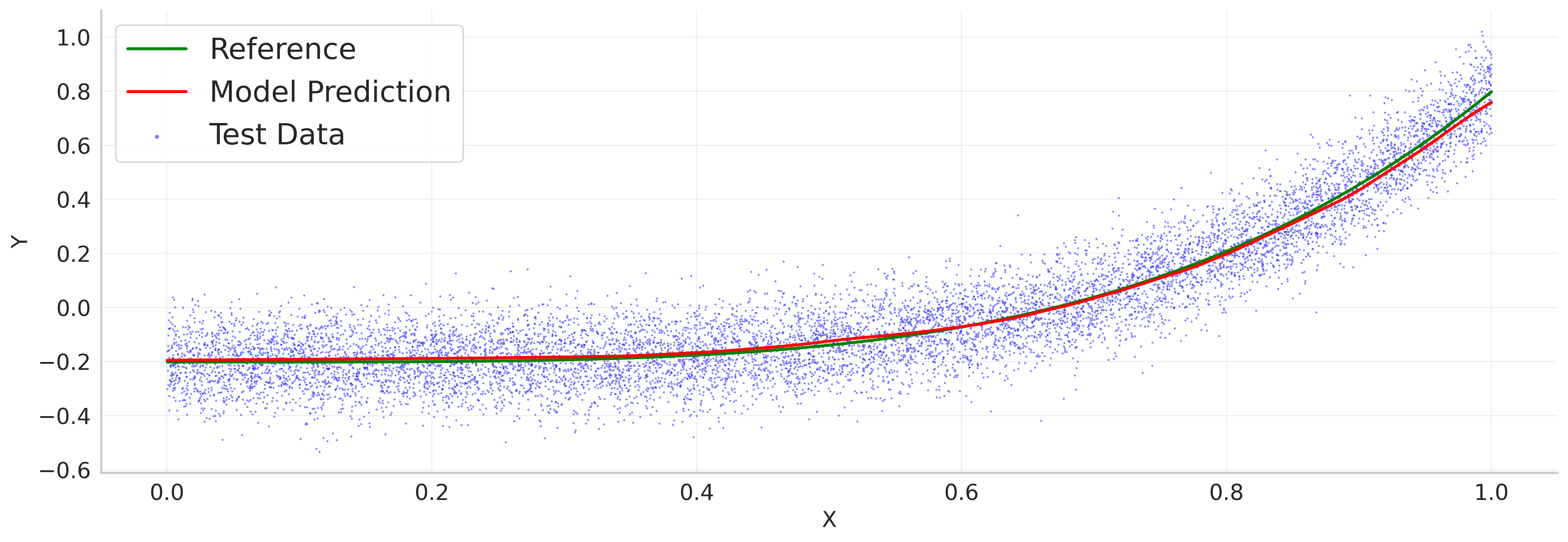}
    \caption{Effect 2}
    \label{fig:ablation_tab_effects_2} 
  \end{subfigure}
  \begin{center}
  \begin{subfigure}[b]{0.5\linewidth}
    \centering
    \includegraphics[width=0.85\linewidth]{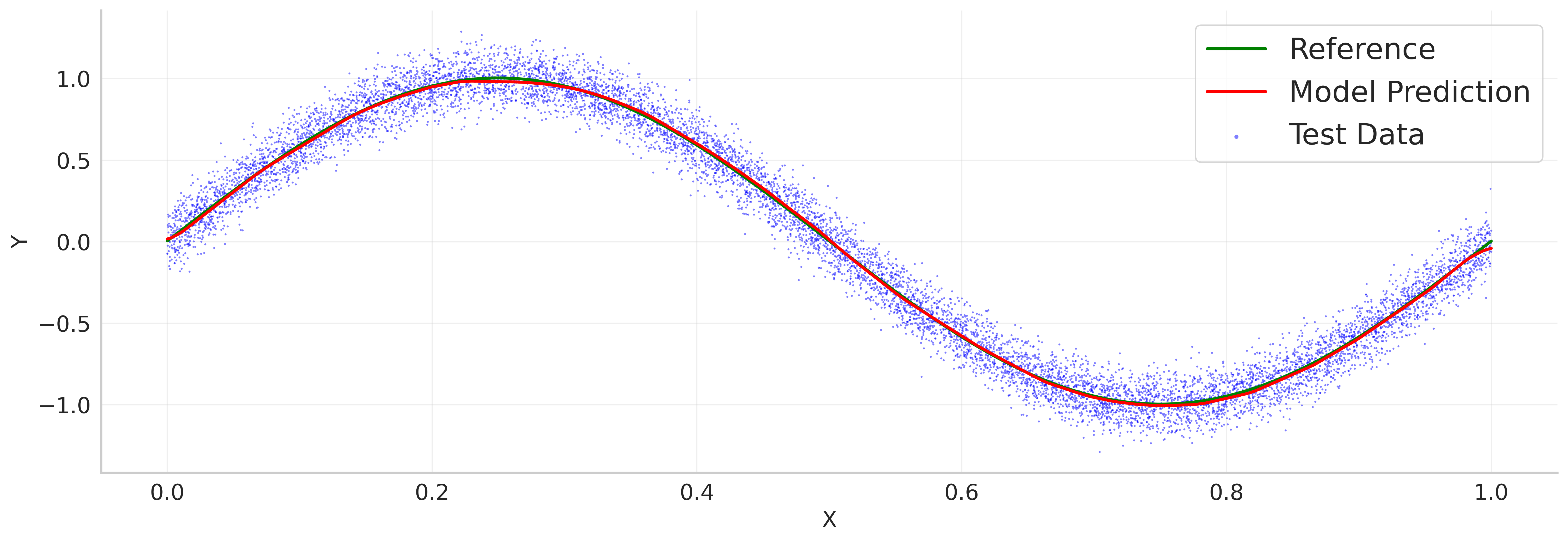}
    \caption{Effect 3}
    \label{fig:ablation_tab_effects_3} 
  \end{subfigure}
  \end{center}
\caption{The NAIM models are able to almost perfectly recover the effects of numerical covariates even if an additional image-covariate is present. Effect 1 is a simple linear effect of the form $f_1(x) = 2x$, effect 2 is a power function $f_2(x) = x^2$ and effect three has a sinusoidal form $f_3(x) = \sin (2 \pi x)$. This plot shows the discovered numerical effects for the squares-data $\Ds$ with $f_{img}(x) = 2x^4$ as the image effect of the x-coordinate of the center of the white square.}
  \label{fig:ablation_tab_effects}
\end{figure}

Regarding the identifiability issue of numerical effects, we also find that the NAIM framework can correctly and accurately identify the effect of several numerical covariates when combined with an image-effect (Figure \ref{fig:ablation_tab_effects} and Table \ref{table:nuemrical_fit} in the supplementary material). Note that for plotting purposes, we zero-center the training data and the model's prediction.


\subsubsection{Image Effects}

\begin{figure}[h]
    \centering
    \includegraphics[width=.95\textwidth]{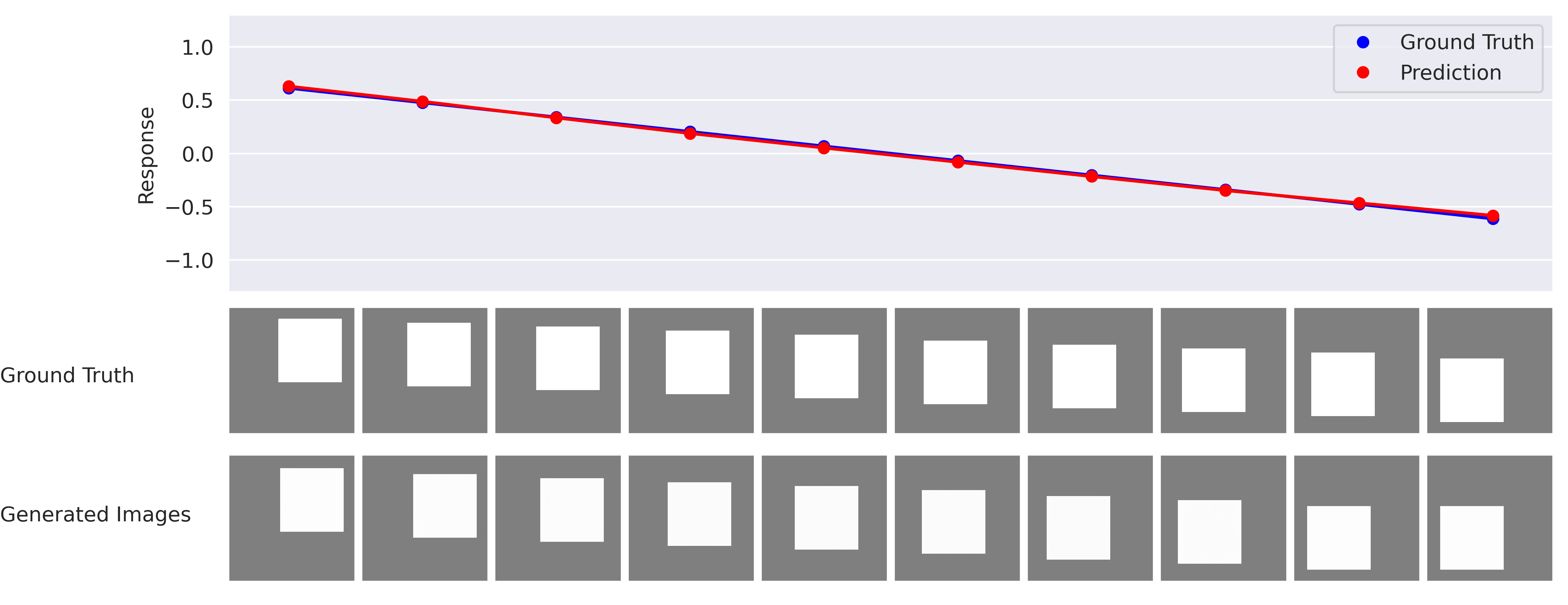}
    \caption{Visualizing that the NAIM framework can almost perfectly recover a linear effect of the {x-coordinate} of a white square on grey background in the form $f_{img}(x) = 2x$.}
    \label{fig:ablation_squares_1}
\end{figure}

\begin{figure}[h]
    \centering
    \includegraphics[width=.95\textwidth]{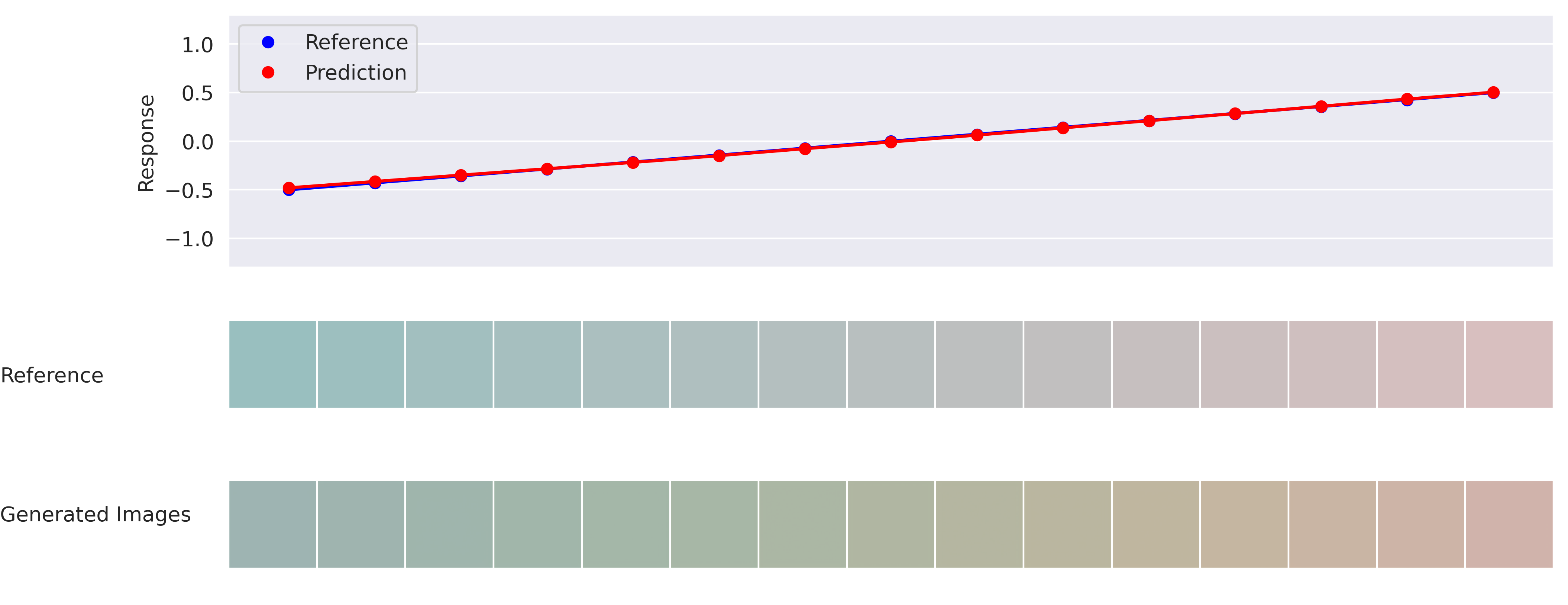}
    \caption{Visualizing that the NAIM framework can almost perfectly recover a linear effect of the {red value} of monochromatic RGB-image in the form $f_{img}(x) = 2x$.}
    \label{fig:ablation_colors_1}
\end{figure}

\begin{figure}[h]
    \centering
    \includegraphics[width=.95\textwidth]{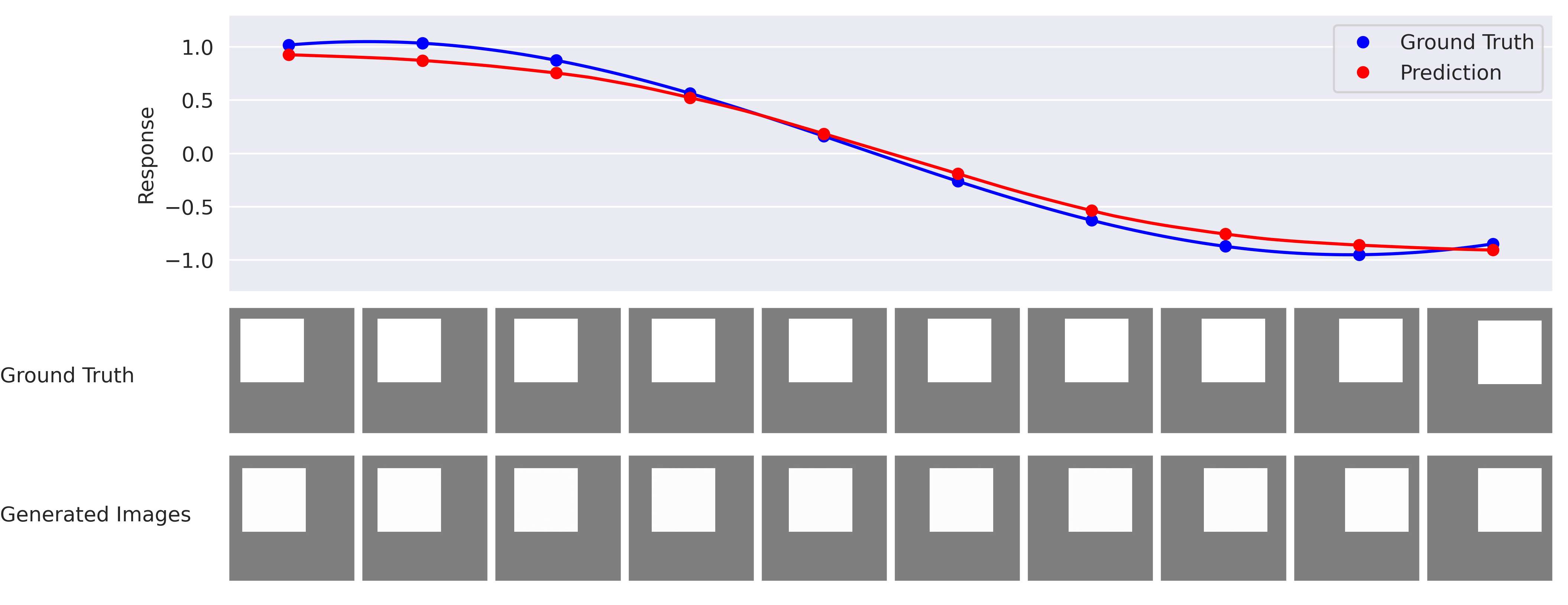}
    \caption{Visualizing that the NAIM-Framework can recover a sinusoidal effect in the form $f_{img}(x) = \sin(2 \pi x)$ where $x$ is of the {x-coordinate} of a white square on grey background.}
    \label{fig:ablation_squares_2}
\end{figure}

\begin{figure}[h]
    \centering
    \includegraphics[width=.95\textwidth]{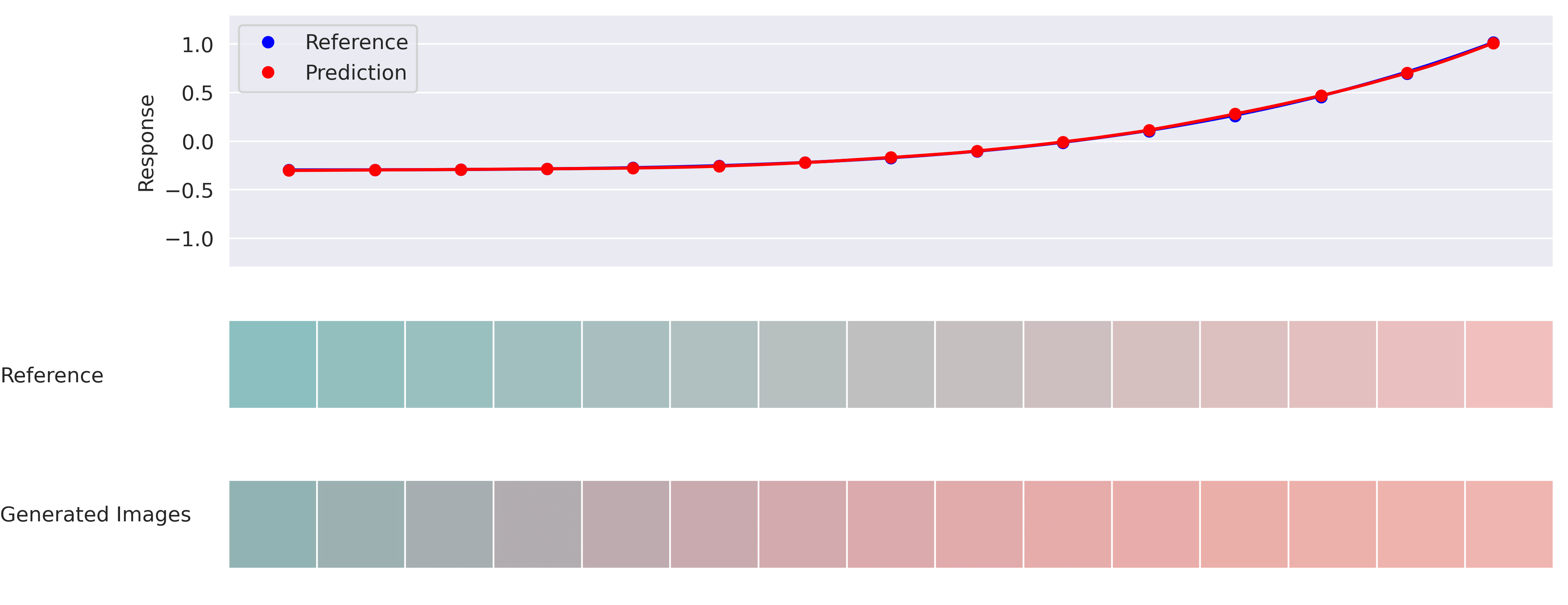}
    \caption{Visualizing that the NAIM-Framework can almost perfectly recover an effect of the form $f_{img}(x) = 2x^4$ where $x$ corresponds to the {red value} of monochromatic RGB-images.}
    \label{fig:ablation_colors_2}
\end{figure}

Beyond the numerical covariates, we find that the NAIM model is capable of reliably identifying different image effects among several additive effects. More precisely, we validate the capabilites in recovering the image-effects by choosing 1,000 pairs of images $\mathcal{D}_{pairs} = \left\{ \left( \bm{x}_{img}^{i_1}, \bm{x}_{img}^{i_2} \right) \right\}_{i=1}^{1000} $ from the test set of the squares-data $\Ds$ and the colors data $\Dc$, followed by interpolating between the images with 1,000 interpolation steps for each pair. On the one hand, we interpolate with the NAIM between the two elements of the $\left( \bm{x}_{img}^{i_1}, \bm{x}_{img}^{i_2} \right)$ tuples, which yields a sequence of semantic subcodes and stochastic subcodes (see section \ref{section:Image_effect_interpretation}).
On the other hand, we linearly interpolate between the ground-truth features $\left(\bm{\Phi}(\bm{x}_{img}^{i_1}), \bm{\Phi}(\bm{x}_{img}^{i_2}) \right)$ of the images where $\bm{\Phi} = \bm{\Phi}_{xval}$ for $\Ds$ and  $\bm{\Phi} = \bm{\Phi}_{red}$ for $\Dc$. This means we linearly interpolate the position of the white square and the red value of the image, respectively. We then compare the model's prediction based on the model's interpolation to the true effect of the linear interpolation of the simple ground-truth features using the mean-squared-error ($MSE$) and the coefficient of determination ($R^2$).

Our experimental results show that the NAIM is able to recover and identify the true image-effects reliably by interpolating between individual samples. 

\begin{table}[h]
\centering
\begin{tabular}{|l|l|l|l|}
\hline
\multicolumn{4}{|c|}{Squares Data} \\
\hline
Type of image effect & $f_{img}(x)$                                  & $MSE$                 & $R^2$ \\ \hline
Linear               & $2x$                                          & $1.82 \cdot 10^{-4}$ & $0.996$ \\
Power function       & $2x^4$                                        & $3.89 \cdot 10^{-3}$ & $0.882$ \\
Sinusoidal           & $\sin( 2 \pi x)$                              & $1.33 \cdot 10^{-2}$ & $0.952$ \\ \hline
\end{tabular}

\vspace{1cm}

\begin{tabular}{|l|l|l|l|}
\hline
\multicolumn{4}{|c|}{Colors Data} \\
\hline
Type of image effect & $f_{img}(x)$                                  & $MSE$                   & $R^2$ \\ \hline
Linear               & $2x$                                          & $1.70 \cdot 10^{-4}$   & $0.968$ \\
Power function       & $2x^4$                                        & $2.57 \cdot 10^{-4}$   & $0.933$   \\
Sinusoidal           & $\sin( 2 \pi x)$                              & $1.94 \cdot 10^{-4}$      & $0.968$   \\ \hline
\end{tabular}
\caption{For benchmarking the capabilities of the NAIM framework in identifying image effect, we compute average $MSE$ and $R^2$ scores for 1000 randomly selected image pairs from the test set of the squares data and the colors data. This allows to assess how closely the NAIM can recover the true effect of an image covariate. Here, the image effect is computed as the function $f_{img}$ of the x-value of the centroid of a white square for the squares data on grey background and the red value of the image for the colors data.}
\label{table:image_effects}
\end{table}

Figure \ref{fig:ablation_squares_1}, \ref{fig:ablation_colors_1}, \ref{fig:ablation_squares_2} and \ref{fig:ablation_colors_2} demonstrate that the model can accurately capture and depict image effects.

The interpolation of the latent codes for the square-data corresponds very closely to linear interpolation of the white-square positions (Figure \ref{fig:ablation_squares_1} and Figure \ref{fig:ablation_squares_2}). While the correct linear effect is almost perfectly identified, the decoded images slightly deviate from the reference in case of the colors-data, which is to be expected since semantic interpolation in this case is much less obvious and certainly not unique (Figure \ref{fig:ablation_colors_1} and Figure \ref{fig:ablation_colors_2}).

Note that this ablation study is about an inquiry into assessing how well the model can rediscover a specific data-generating process. We want to stress, that all plots of the effects still allow for a perfect explanation of the model itself, although the model might not be capable of capturing the true data-generating process correctly. Notably, when the model diverges from the real data generation process, its explainability becomes vital in identifying and acknowledging these discrepancies. Thus, the proposed method retains its relevance, even if it does not aim to pinpoint the exact effects.

\section{Application}
\subsection{Data}

To assess the effectiveness of the proposed approach for real-world data, we conduct a case study utilizing two Airbnb datasets pertaining to the cities of Cape Town and New York\footnote{Airbnb is an online platform that allows individuals to rent out their properties to guests for short-term stays. The data can be found under \url{http://insideAirbnb.com/}.}. These datasets comprise various types of information, including numerical and categorical features, as well as unstructured data in the form of images.
In our analysis, we employ the logarithm of the rental price as the target variable and apply standard normalization to ensure consistency.
The available information includes, among further covariates, the number of bedrooms, bathrooms and beds, as well as the average review rating, the total number of reviews and the exact location of the property in the form of longitude and latitude. 
We preprocess the tabular features by standardizing the numerical covariates and using median imputing for missing values. 

Regarding the images, we use the host's profile pictures that are uploaded by the hosts themselves and that therefore may contain noise. These images typically portray the actual person, with the composition strongly varying depending on the preferences of the users. Therefore, we align the images of the hosts in order to obtain more consistent image sections and head poses, using the alignment procedure employed by \citet{preechakul2022diffusion}. Thereby we also discard all images that do not depict human faces and thus we ignore all datapoints where no picture of a human is available. It is also important to note that the variance within the images is still quite substantial and that sometimes an image section showing one person out of a group within the profile picture is selected. 
By utilizing this rich and diverse dataset, we are able to comprehensively analyze the impact of various image characteristics on rental prices in the Airbnb domain. 
Through our approach of semantic image interpolation, we seek to uncover the underlying patterns and connections between image semantics and the pricing of Airbnb rentals, providing valuable insights into the factors that drive changes in rental prices based on image attributes.
The spatial distribution of the rental offers can be seen in Figure \ref{fig:capetown_city} in the supplemental material.
\begin{table}
\centering
\caption{Performance comparison for comparable interpretable additive models. NAM \citep{agarwal2021neural}, EBM \citep{nori2019interpretml} and NodeGAM \citep{chang2021node} are all fit without the image effect as the architectures of EBM and NodeGAM do not allow to incorporate high dimensional data.}
\begin{tabular}{lcc}
\toprule
\multicolumn{1}{l}{\multirow{2}{*}{Model}} & \multicolumn{2}{c}{$R^2$-score}          \\ \cline{2-3} 
& Cape Town & New York \\ 
\midrule
NAM  & 0.4114 & 0.4657   \\
NAIM  & \textbf{0.4778} & \textbf{0.5008}   \\
EBM  & 0.4393   & 0.4783   \\
NodeGAM    & 0.4241  & 0.4815   \\ \bottomrule
\end{tabular}
\label{tab:benchmark}
\end{table}

For our analysis, we fit the proposed Neural Additive Image Model (NAIM) including the numerical covariates "room type", "accommodates", "bedrooms", "minimum nights", "number of reviews", "review scores value" and the host image as an image effect. See part \ref{apx:CovariateExplanaition} in the supplementary material for more details on the covariates. Compared to equivalently  interpretable models that only use those numerical covariates, including the image effect increases the coefficient of determination from $R^2 = 0.411$ to $R^2 = 0.478$ for the Airbnb data from Cape Town and from $R^2 = 0.465$ to $R^2 = 0.508$ for New York (see Table \ref{tab:benchmark}). While NodeGAM \citep{chang2021node} and EBM \citep{nori2019interpretml} outperform the simple NAM \citep{agarwal2021neural}, NAIM outperforms all models for both datasets by also incorporating the image effect.

\subsection{Interpretability}

\begin{figure}[ht]
  \begin{subfigure}[b]{0.5\linewidth}
    \centering
    \includegraphics[width=0.85\linewidth]{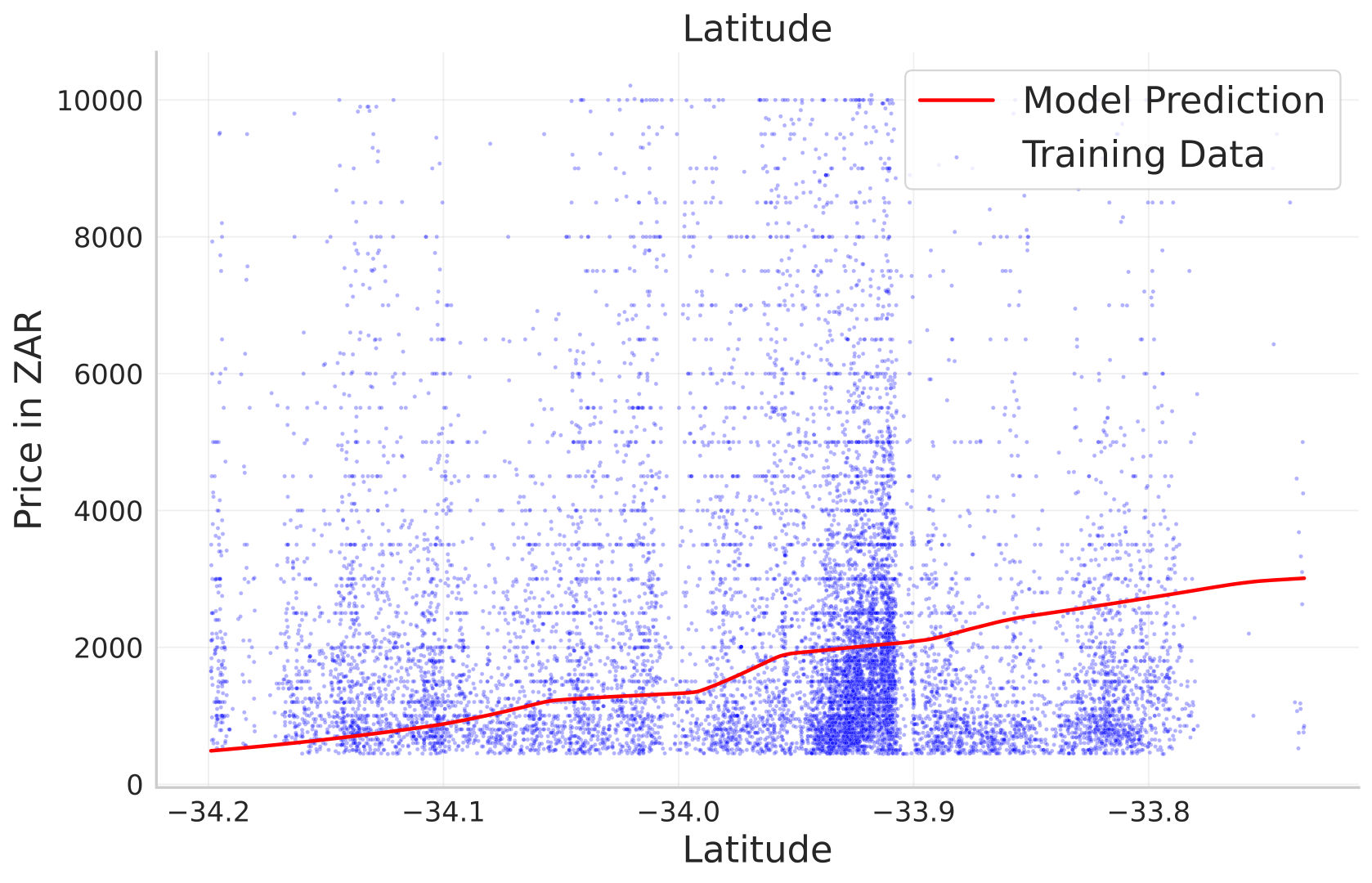}\label{fig:gray hair}
    \caption{Latitude} 
    \label{fig7:Latitude} 
  \end{subfigure}
  \begin{subfigure}[b]{0.5\linewidth}
    \centering
    \includegraphics[width=0.85\linewidth]{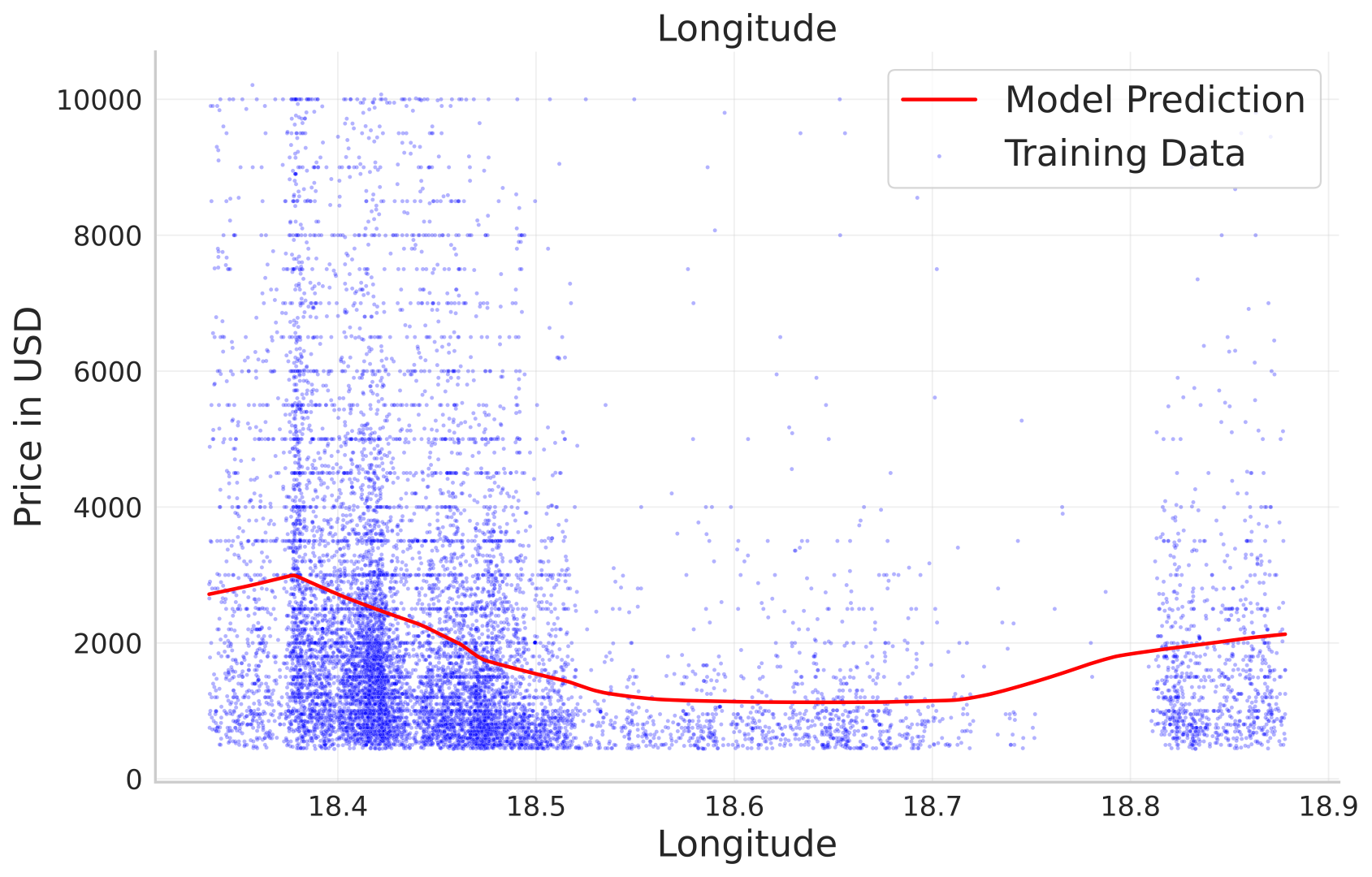}\label{fig:Longitude}
    \caption{Longitude} 
    \label{fig7:Longitude} 
  \end{subfigure} 
  
  \begin{subfigure}[b]{0.5\linewidth}
    \centering
    \includegraphics[width=0.85\linewidth]{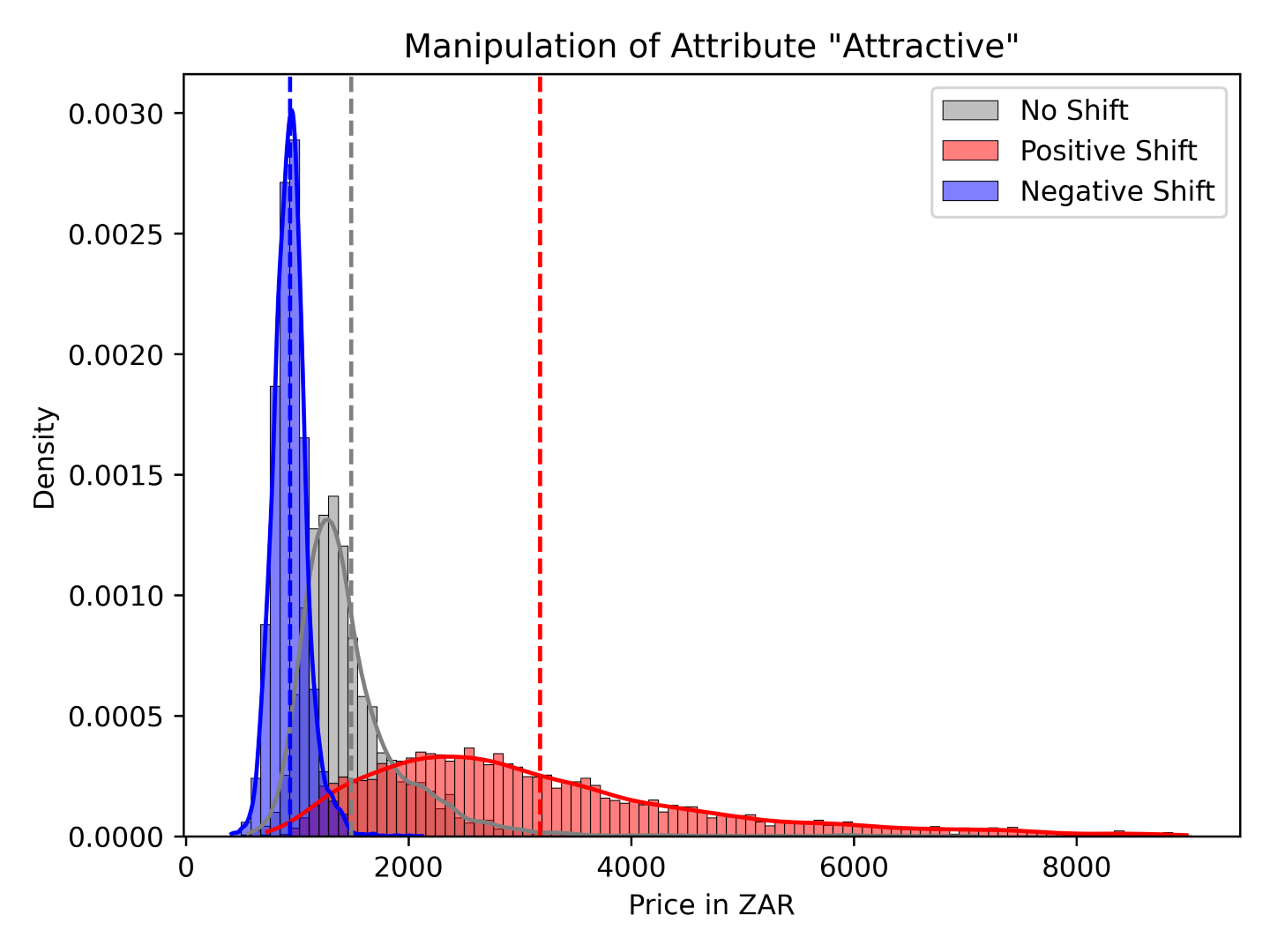}\label{fig:global_Att}
    \caption{Attractiveness} 
    \label{fig7:attractiveness} 
    \vspace{4ex}
  \end{subfigure}
  \begin{subfigure}[b]{0.5\linewidth}
    \centering
    \includegraphics[width=0.85\linewidth]{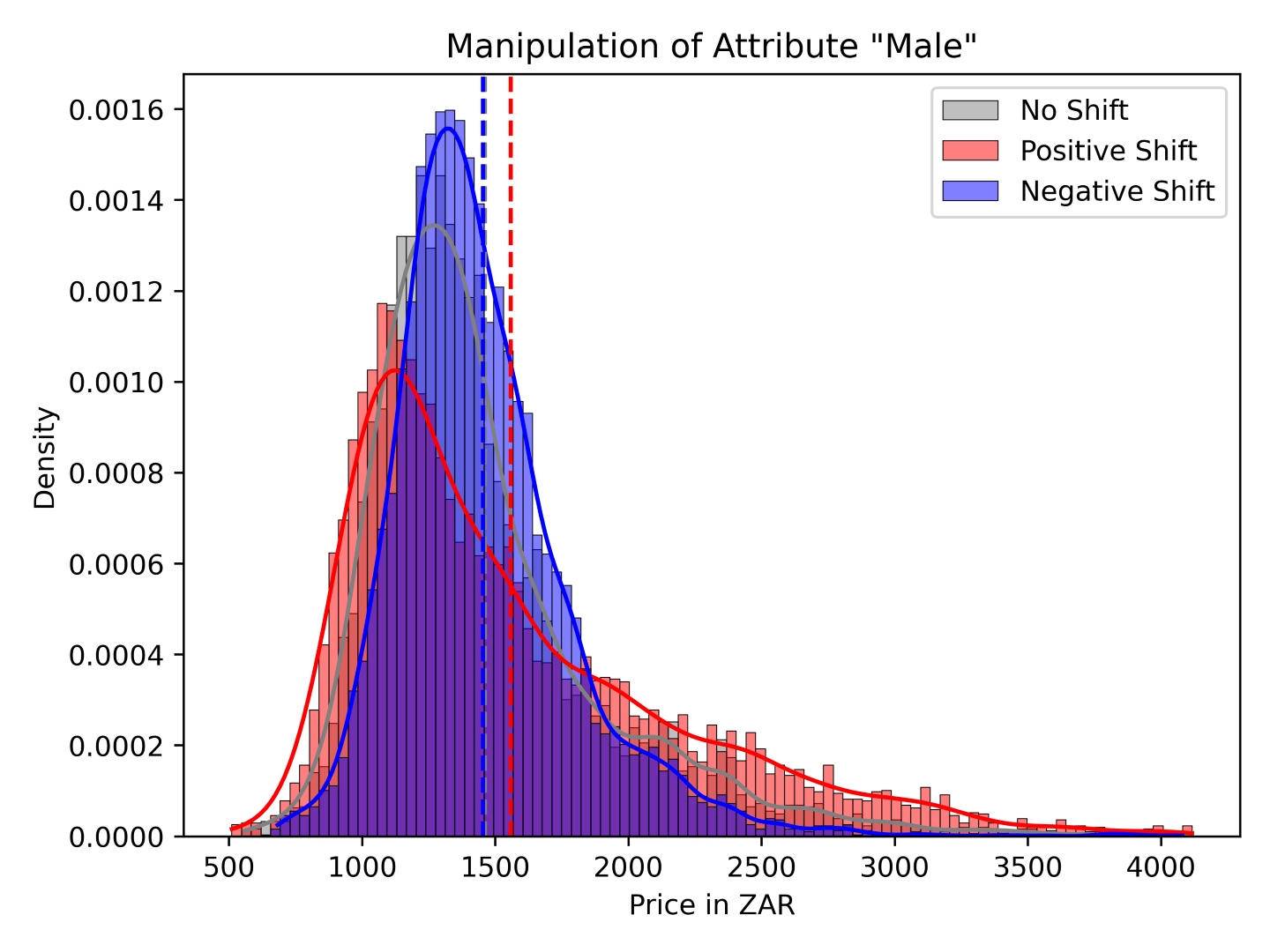}\label{fig:f2}
    \caption{Gender} 
    \label{fig7:gender} 
    \vspace{4ex}
  \end{subfigure} 

  \caption{Global interpretation of effects in a Neural Additive Image Model. The effects of numerical covariates can by simply visualized by plotting the corresponding shape functions. 
  The global effect of a certain image features can by analyzed through manipulating all images in the data set analogously and visualizing the shift of the predictive distribution. The grey distribution depicts the original distribution of the predictions for the entire dataset without any manipulation while blue distribution depicts a negative shift and the red distribution depicts a positive shift on the semantic subcode for the chosen feature respectively. We further visualize the means of the distributions with dashed vertical lines.} 
  
  \label{fig:global_effects}
\end{figure}

First, we demonstrate the global interpretability of the tabular features. Similar to GAMs \citep{hastie2017generalized}, NAMs \citep{agarwal2021neural}, EBMs \citep{nori2019interpretml} or NodeGAMs \citep{chang2021node}, the individual feature effects can be visualized without further adjustments. Figures \ref{fig7:Latitude} and \ref{fig7:Longitude} demonstrate the effects of some of the variables for New York.
Global image effects, however, are not as easily visualizable. By shifting the embeddings of all input images $\{\bm{E}_\phi(\bm{x}_{img}^{(i)}) \}_{i=1 \ldots n}$ by the same vector into a specified direction and analyzing both, the predictive distribution over all unshifted images, as well as the predictive distribution over the shifted images, we can infer global effects of image characteristics. See section \ref{sec:DAEs} for more details on how the vector used to manipulate an attribute is obtained.
Figures \ref{fig7:attractiveness} and \ref{fig7:gender} show the global effects of the abstract features \textit{attractiveness} and \textit{gender} for Cape Town.

The global effects also become apparent when visualizing them on a sample level. This allows, to not only identify biases on a global level, but to also detect biases for predictions based on individual examples, and can thus prevent resulting harm in potential real-world decisions. The identified increase of the price for more attractive hosts (Figure \ref{fig7:attractiveness}) is also identifiable for single hosts. In order not to violate any personal rights we do not depict the real host images for these examples. Instead, we choose examples from the CelebA dataset \citep{liu2015faceattributes}. However, the model training as well as the global shifts are all performed on the real host images of the hosts of Airbnb rentals. For the image encoder and the DDIM, we use the weights provided by \citet{preechakul2022diffusion}.
Figure \ref{fig:chubby_plus} depicts a positive shift  of the host's image into the direction described as "chubby" in the CelebA dataset. The global effects of this feature can be found in the supplementary material.

\begin{figure}
    \centering
    \includegraphics[width=.95\textwidth]{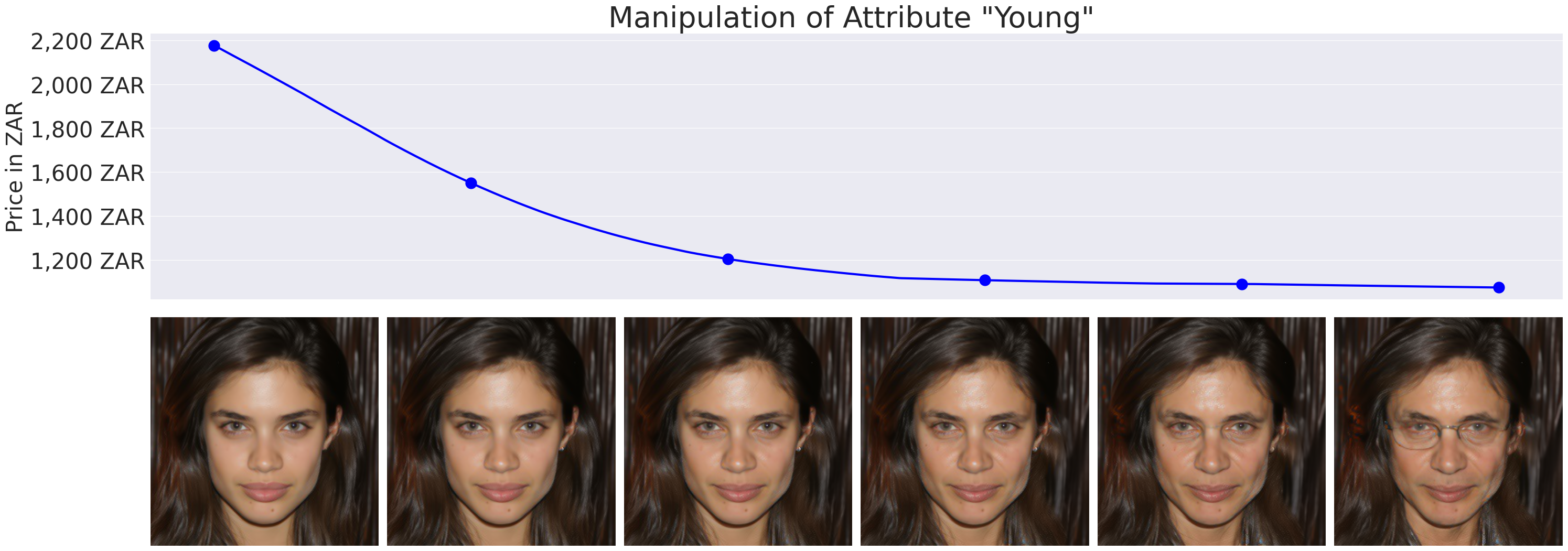}
    \caption{Effect of the feature \textit{age} on the expected rental price. Through (semi) continuous interpolation, an effect trend can be visualized.}
    \label{fig:age}
\end{figure}

\begin{figure}
    \centering
    \includegraphics[width=.95\textwidth]{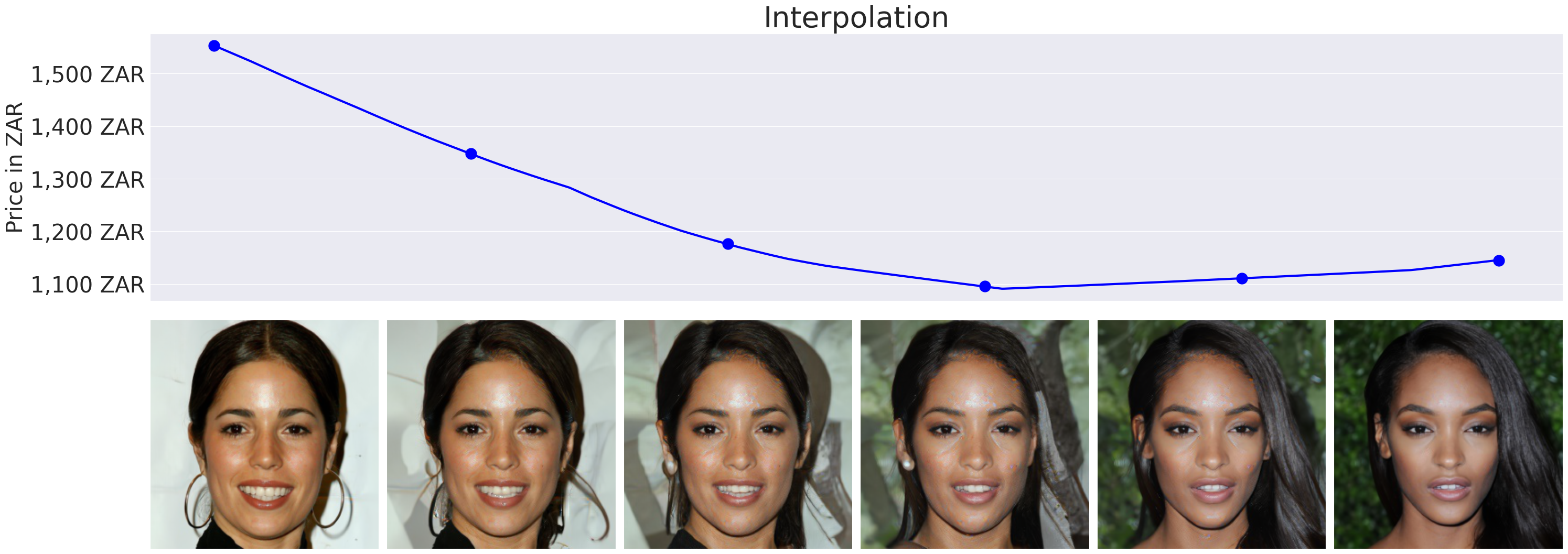}
    \caption{Visualizing the effect of interpolation between two samples on the expected rental price in Cape Town.}
    \label{fig:interpolation2}
\end{figure}

Figures \ref{fig:interpolation_skincolor} and \ref{fig:interpolation2} provide a means to visually depict the impact on the expected rental price resulting from an interpolation between the two images shown at the far left and far right. This interpolation effectively captures various attributes, e.g., including skin color and hair color and thus also depict attribute interactions. Since no explicit attribute manipulation is needed, interpolation also allows to investigate various features in a very flexible way, depending on the concrete use-case. 

\section{Broader Impact: On the Importance of Interpretability}
Interpretable machine learning models have become increasingly important in real-world applications where transparency and fairness are critical. In particular, additive models have gained popularity due to their ability to provide interpretable and accurate predictions by decomposing the effects of input features into additive components \citep{jo2022neural, chen2022monotonic, moslehi2022interpretable, siems2023interpretable}. However, in the case of image data, the interpretability of additive models has been limited due to the complex and highly nonlinear nature of image features.
Especially in the context of fairness and discriminatory effects, image interpretability is a crucial topic in the field of machine learning:

\textbf{(I)} Images can contain highly personal and very sensitive information. In many domains, such as hiring or social media they provide very detailed information on features such as gender, age, body weight or race of a person.

\textbf{(II)} Crucial information is not provided via simple numerical values, but often hidden within partly impenetrable image semantics (e.g. \citep{pingitore1994bias}). The highly complex and somewhat opaque nature of image features makes it very difficult to assess if a given algorithm that relies on image features does actually use morally problematic features for its predictions.

\textbf{(III)} Full intelligibility of image effects allows researchers to interrogate deep learning model behaviors and mitigate system malfunctions.

\textbf{(IV)} As machine learning models become more prevalent in real-world applications, such as job applications, finance or healthcare, the potential for biased decision-making increases, which can be identified with fully intelligible models.

\textbf{(V)} Leveraging complete image intelligibility can be particularly beneficial in the field of medical imaging analysis. The presented approach could uncover valuable insights that may aid in the early detection of diseases and enhance diagnostic quality. 

%
%

\begin{figure}
  \centering
{\includegraphics[width=0.37\textwidth]{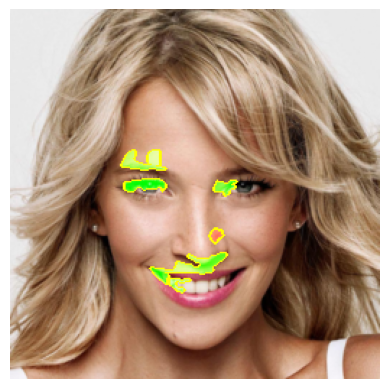}\label{fig:global_Att}}
  \hfill
{\includegraphics[width=0.37\textwidth]{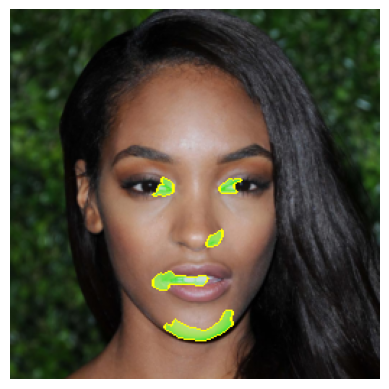}\label{fig:f2}}
  \caption{Post-hoc analysis of pixel importance using LIME \citep{ribeiro2016should} for a regression model predicting expected rental prices based on host images. For data protection reasons, we depict images from the CelebA dataset, instead of using the original images of private persons.}
  \label{fig:LIME}
\end{figure}

While post-hoc methods, such as LIME \citep{ribeiro2016should} or Shapely values \citep{shapley1953quota}, are capable to identify important pixel regions, the interpretation of these pixel regions is often not sufficient for generating deep understanding of humanly interpretable image effects and does not allow for a quantification of the image effect. Figure \ref{fig:LIME} demonstrates, that while relevant pixel areas are identified, they lack any form of easy interpretability. This is especially true for continuous dependent variables, $y\in \mathbb{R}$, since it is not easily interpretable whether the identified pixel regions have positive or negative effects on $y$.  Our presented model thus proposes to rethink the impact of images on numerical features as a form of continuous effect, given that visual attributes are rarely discretely categorizable. For instance, effects that can be extracted from images, such as a person's hair color, should not be treated as a discrete feature, as there are infinitely many colors between any two hair colors.

By ensuring that machine learning models are fully interpretable and transparent, we can identify, understand and mitigate these biases, promoting fairness and equity in decision-making.

\section{Conclusion}
In recent years, there has been a growing demand for interpretability in machine learning models, particularly in models that deal with highly complex data. While deep learning approaches have demonstrated high predictive performance, their black-box nature has been a major hurdle for their adoption in fields where interpretability is a key requirement.

To address this challenge, we present a novel approach to account for full interpretability in the context of multi-modal data. We are able to visualize tabular data similarly to \citet{agarwal2021neural} and \citet{chang2021node}. Additionally, we are able to construct interpretable representations of semantically meaningful image effects. Thereby, we could firstly demonstrate global and a-priori feature interpretability as well as global (see Figure \ref{fig:global_effects}) and local (e.g. Figure \ref{fig:age}) post-hoc interpretability of image effects. Through interpolation as well as attribute manipulation, the presented approach offers possibilities for image interpretation in a nearly unlimited number of scenarios.


In conclusion, we present an effective approach that addresses the interpretability challenges encountered in modeling multi-modal data. It achieves this by ensuring interpretability at both the feature and image level. This comprehensive interpretability fosters greater trust and deeper comprehension regarding models handling use-cases with multi-model data, thereby promoting wider acceptance in domains where understanding the model's inner workings is essential.

\section{Limitations and Discussion}
The current work offers intelligibility for image effects in multi-modal data problems.

For the NAIM model, we establish identifiability by using feature dropout, as presented by \citet{agarwal2021neural} and utilize it to not only drop-out tabular effects, but also the effect of image covariates. However, when assuming that especially high-dimensional image effects can be highly correlated with other features, it might be reasonable to investigate including more recent developments regarding identifiablity in the realm of multi-model data, such as presented in \citet{rugamer2022semi, siems2024curve}.
Those developments might be especially beneficial for our approach when exploring the possibility of including image-image interactions or even interactions between metric features and images. 


The presented case study is focused on feature attributes from faces and can thus leverage pre-trained and identified attributes from previous work \citep{preechakul2022diffusion}. The introduced approach, however, is much more general and can be applied to a practically unlimited variety of different images datasets by simply pre-training the autoencoder on the respective  data. Furthermore, future work could even investigate the possibility of integrating unstructured covariates other than images, for instance texts.

\section{Ethics Statement}




Throughout this work we present and showcase a method that can be used to analyze the impact of attributes of Airbnb hosts' faces on Airbnb prices. Therefore, our method can be used to investigate effects of abstract image attributes such as \textit{attractiveness}, \textit{gender} or \textit{age} that our model might have learned. Thus, our methodology also allows for analyses of biases regarding those features.
However, we want to emphasize that appropriately analysing the structure of biases in the used datasets is beyond the scope of this work, in which we aim at introducing and showcasing the method itself. 
Nevertheless, we want to emphasize that the presented method can indeed have profound societal value in future work, as it can reveal potentially discriminatory effects that can only be identified through interpretable visual analysis.

We train all our models on the true images that are provided by Airbnb directly. However, in this manuscript we demonstrate the visual effects on the CelebA dataset due to privacy concerns. 

\bibliography{bib.bib}

\newpage

\appendix
\section{Supplemental material for Neural Additive Image Models}

\subsection{Proof of Theorem 1}

{\bf{Proof} of \bf{Theorem 1}}{
\label{proof_theorem}
The result follows directly from the definition of total derivatives. In particular, one has 
\begin{equation*}
    h \left(\lambda \bm{z} + (1-\lambda) \tilde{\bm{z}} \right) = h \left(\bm{z} + (1-\lambda) (\tilde{\bm{z}} - \bf{z}) \right) = h(\bm{z}) + D_{\bf{z}}h ((1-\lambda) (\tilde{\bm{z}} - \bm{z})) + (1 - \lambda)R_1,
\end{equation*} 
where $R_1 = \bm{o}(\lvert \lvert \bm{z} - \tilde{\bm{z}}  \lvert \lvert)$. Using that $h( \tilde{\bm{z}}) = h(\bm{z} + (\tilde{\bm{z}} - \bm{z})) = h(\bm{z}) +D_{\bf{z}}h(\tilde{\bm{z}} - \bm{z}) + R_2$, where $R_2 = \bm{o}(\lvert \lvert \bm{z} - \tilde{\bm{z}}  \lvert \lvert)$ and setting $R \vcentcolon = R_1 - R_2$, one further obtains 

\begin{equation*}
\begin{split}
h (\lambda \bm{z} + (1-\lambda) \tilde{\bm{z}} ) = \lambda h(\bm{z}) + (1 - \lambda) \left(h(\bm{z}) + D_{\bf{z}}h (\tilde{\bm{z}} - \bm{z}) \right) + (1-\lambda)R_1 = \\ \lambda h(\bm{z}) + (1 - \lambda) h( \tilde{\bm{z}}) + (1-\lambda)(R_1 - R_2) = \lambda h(\bm{z}) + (1 - \lambda) h( \tilde{\bm{z}}) + (1-\lambda)R, 
\end{split}
\end{equation*}
where clearly also $R = \bm{o}(\lvert \lvert \bm{z} - \tilde{\bm{z}}  \lvert \lvert)$.
}

\newpage
\subsection{Additional results for the experiments with synthetic data}
\label{appendix:synthetic_data}

\begin{figure}[h]
    \centering
    \includegraphics[width=.95\textwidth]{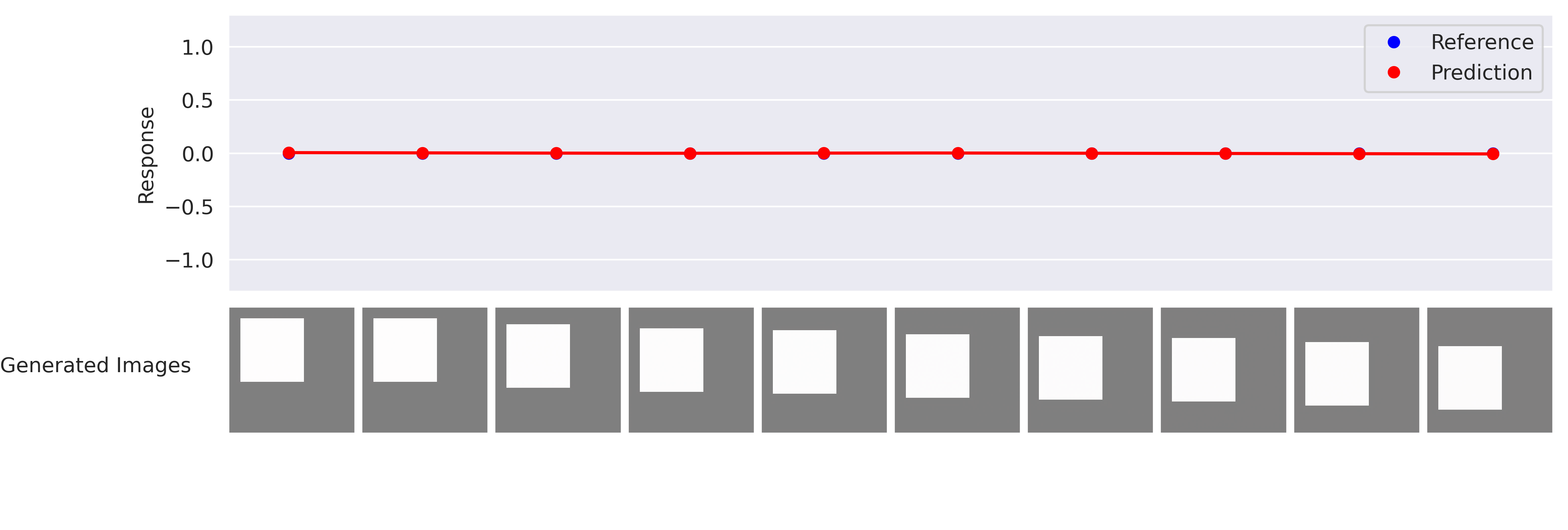}
    \caption{Visualizing that the NAIM-Framework can almost perfectly recover a linear effect of the {x-coordinate} of a white square on grey background in the form $f_{img}(x) = 2x$. Note that in this case, the effect is expected to be zero since the x-coordiante stays the same.}
    \label{fig:ablation_squares_1_appendix}
\end{figure}

\begin{figure}[h]
    \centering
    \includegraphics[width=.95\textwidth]{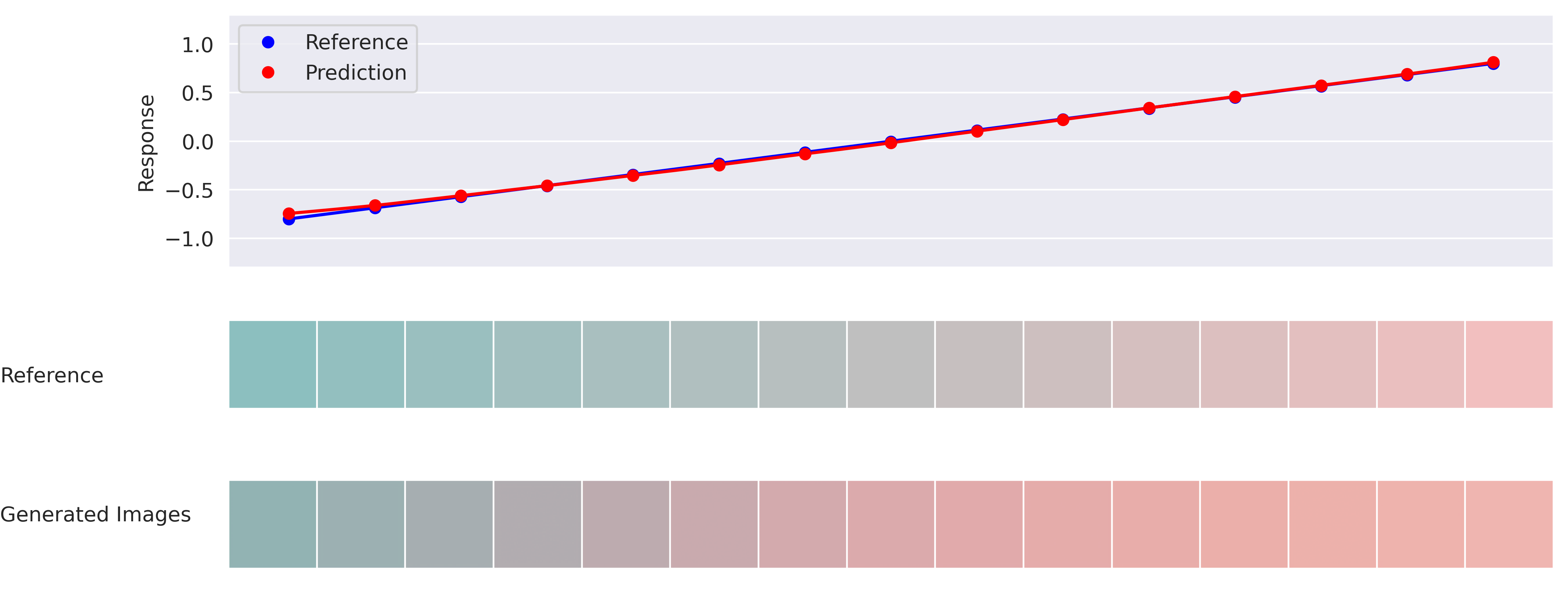}
    \caption{Visualizing that the NAIM-Framework can almost perfectly recover a linear effect of the {red value} of monochromatic RGB-image in the form $f_{img}(x) = 2x$.}
    \label{fig:ablation_colors_1_appendix}
\end{figure}

\begin{figure}[h]
    \centering
    \includegraphics[width=.95\textwidth]{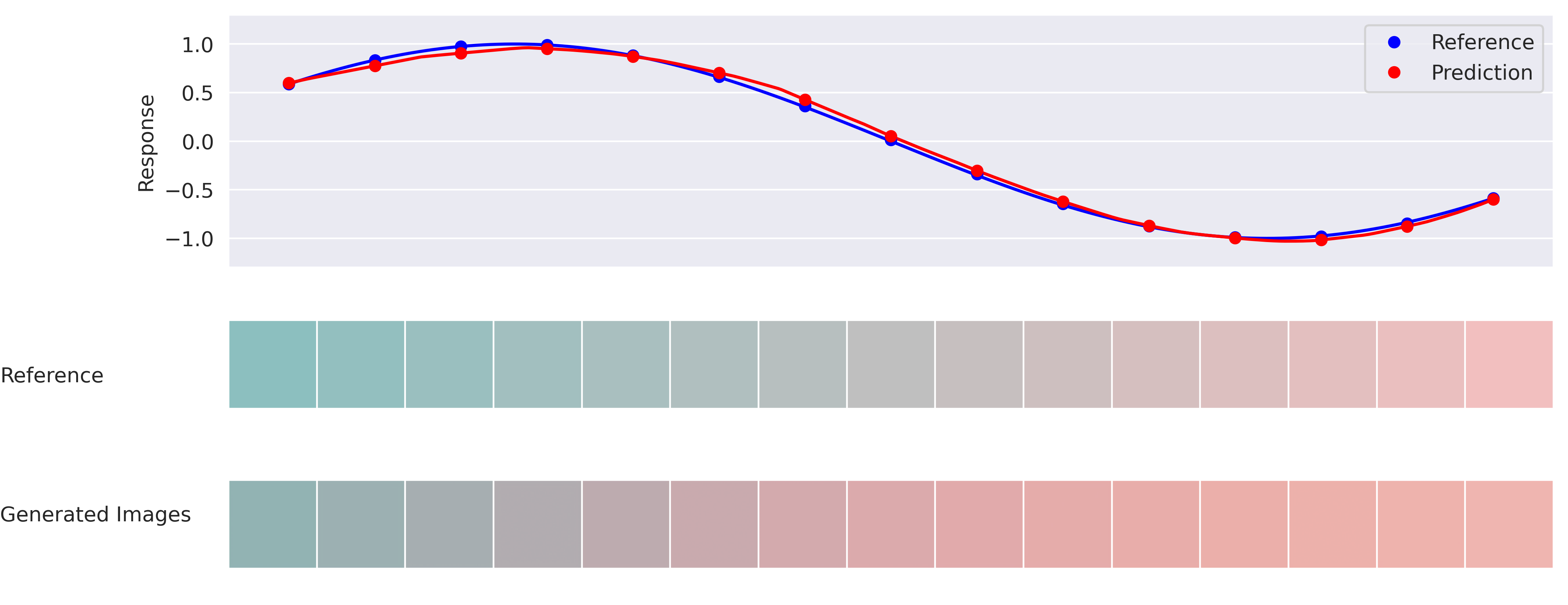}
    \caption{Visualizing that the NAIM-Framework can recover a sinusoidal effect of the {red value} of monochromatic RGB-image in the form $f_{img}(x) = \sin ( 2 \pi x)$.}
    \label{fig:ablation_colors_1_appendix}
\end{figure}

\clearpage
\begin{table}[h]
\centering
\begin{tabular}{|c|c|c|c|}
\hline
\multicolumn{4}{|c|}{Squares Data} \\
\hline
Image effect $f_{img}(x)$  & Numerical effect $f_i$         & $MSE$                              & $R^2$               \\ \hline
$ $                        & $2x$                           & $5.98 \cdot 10^{-4}$              & $0.998$             \\
$2x$                       & $x^2$                          & $4.68 \cdot 10^{-4}$              & $0.994$             \\ 
$ $                        & $\sin( 2 \pi x)$               & $9.67 \cdot 10^{-3}$              & $0.981$             \\ \hline
$ $                        & $2x$                           & $6.85 \cdot 10^{-3}$              & $0.979$             \\
$2x^4$                     & $x^2$                          & $8.15 \cdot 10^{-3}$              & $0.88$             \\
$ $                        & $\sin( 2 \pi x)$               & $1.28 \cdot 10^{-3}$              & $0.998$             \\ \hline
$ $                        & $2x$                           & $6.59 \cdot 10^{-3}$              & $0.980$             \\
$\sin( 2 \pi x)$           & $x^2$                          & $2.46 \cdot 10^{-3}$              & $0.966$             \\
$ $                        & $\sin( 2 \pi x)$               & $5.62 \cdot 10^{-4}$              & $0.999$             \\ \hline
\end{tabular}

\vspace{1cm}

\begin{tabular}{|c|c|c|c|}
\hline
\multicolumn{4}{|c|}{Colors Data} \\
\hline
Numerical effect $f_i$     & Image effect $f_{img}(x)$         & $MSE$                              & $R^2$               \\ \hline
$ $                        & $2x$                              & $0.0938$                          & $0.719$             \\
$2x$                       & $2x^4$                            & $6.59 \cdot 10^{-3}$              & $0.926$             \\ 
$ $                        & $\sin( 2 \pi x)$                  & $0.0396$                          & $0.922$             \\ \hline
$ $                        & $2x$                              & $0.0465$                          & $0.860$             \\
$x^2$                      & $2x^4$                            & $3.71 \cdot 10^{-3}$              & $0.958$             \\
$ $                        & $\sin( 2 \pi x)$                  & $0.0376$                          & $0.925$             \\ \hline
$ $                        & $2x$                              & $5.98 \cdot 10^{-4}$              & $0.998$             \\
$\sin( 2 \pi x)$           & $2x^4$                            & $4.68 \cdot 10^{-4}$              & $0.994$             \\
$ $                        & $\sin( 2 \pi x)$                  & $9.67 \cdot 10^{-3}$              & $0.981$             \\ \hline
\end{tabular}
\caption{Assessing the capability of the NAIM to identify effects of numeric covariates when an additional image covariate is present. The $MSE$ and $R^2$ values are computed between the predicted effect for each $f_i$ and the true value this effect has on a test-set of size 10,000.}
\label{table:nuemrical_fit}
\end{table}
\newpage

\subsection{Additional Manipulation Plots}

\begin{figure}[h]

  \begin{subfigure}[b]{1 \linewidth}
     \centering
     \includegraphics[width=.9\textwidth]{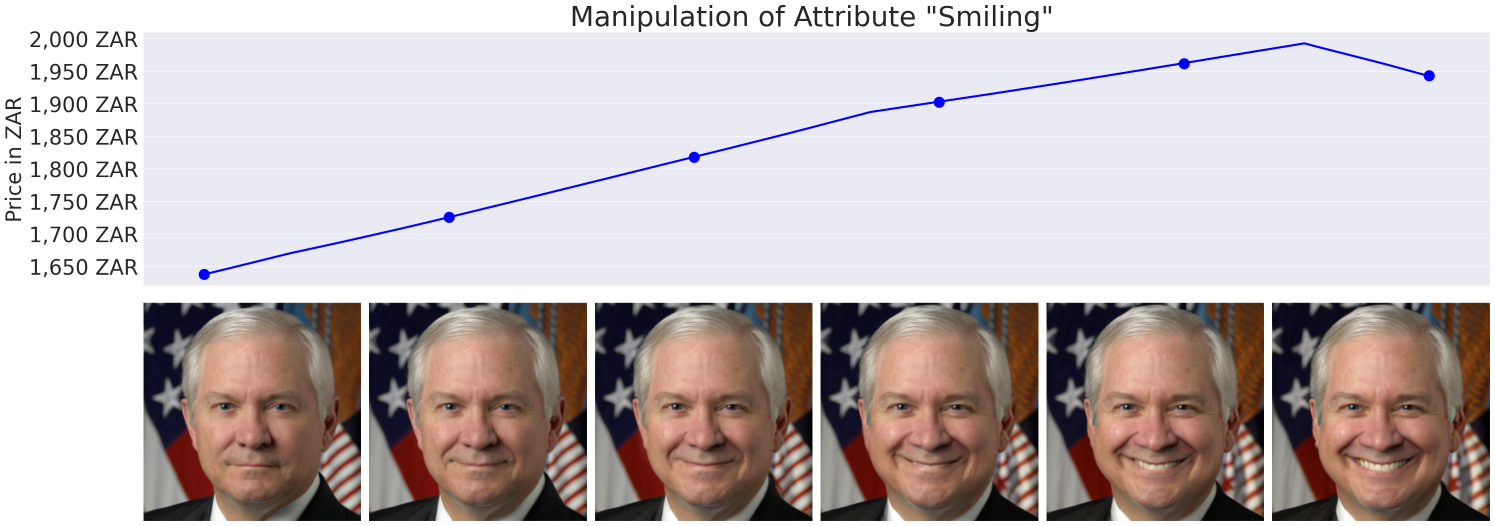}
    \label{fig:interpolation 2}
    \vspace{0.5cm}
  \end{subfigure}
  
\begin{subfigure}[b]{1 \linewidth}
     \centering
    \includegraphics[width=.9\textwidth]{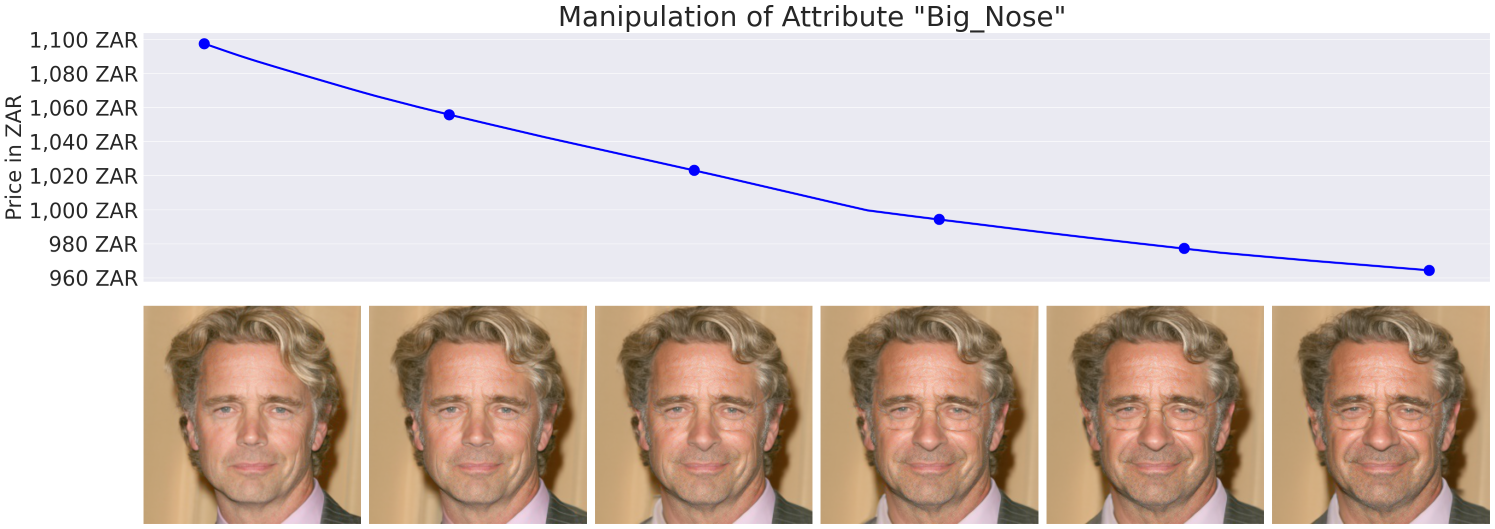}
    \label{fig:interpolation 2}
    \vspace{0.5cm}
    \vspace{0.5cm}
  \end{subfigure}

  \caption{Effect of various facial features on the expected rental price. Through (semi) continuous interpolation, an effect trend can be visualized.}
\end{figure}

\begin{figure}[H]
\begin{subfigure}[b]{1 \linewidth}
     \centering
    \includegraphics[width=.9\textwidth]{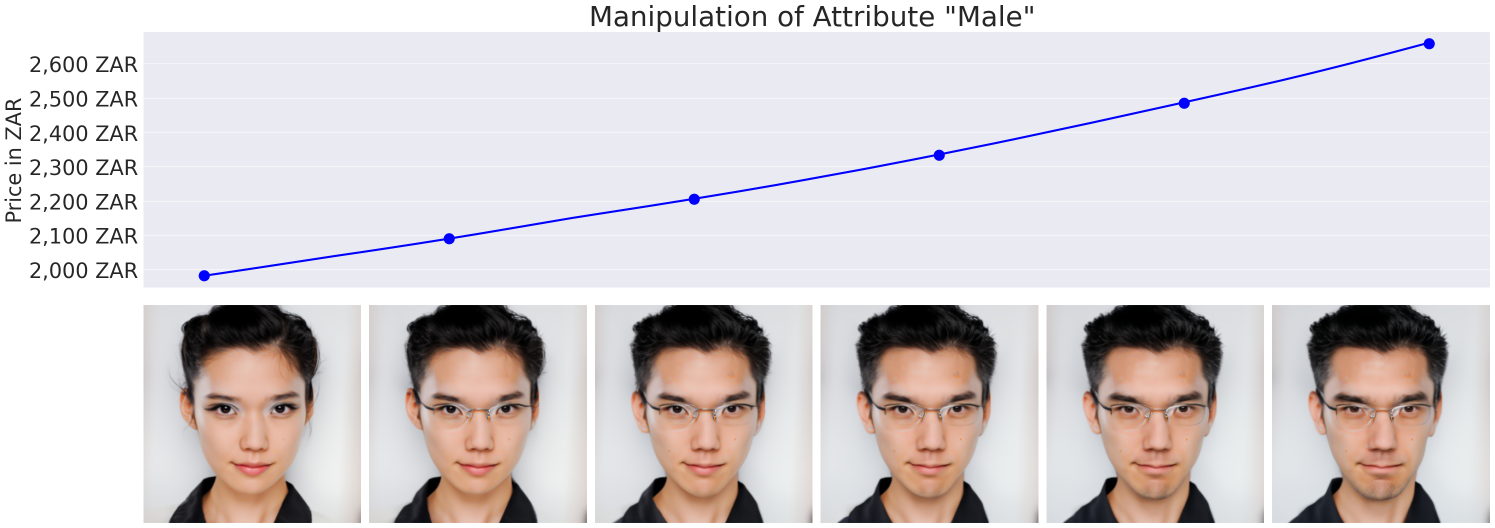}
    \label{fig:interpolation 2}
    \vspace{0.5cm}
  \end{subfigure}

  \begin{subfigure}[b]{1 \linewidth}
    \centering
     \includegraphics[width=.9\textwidth]{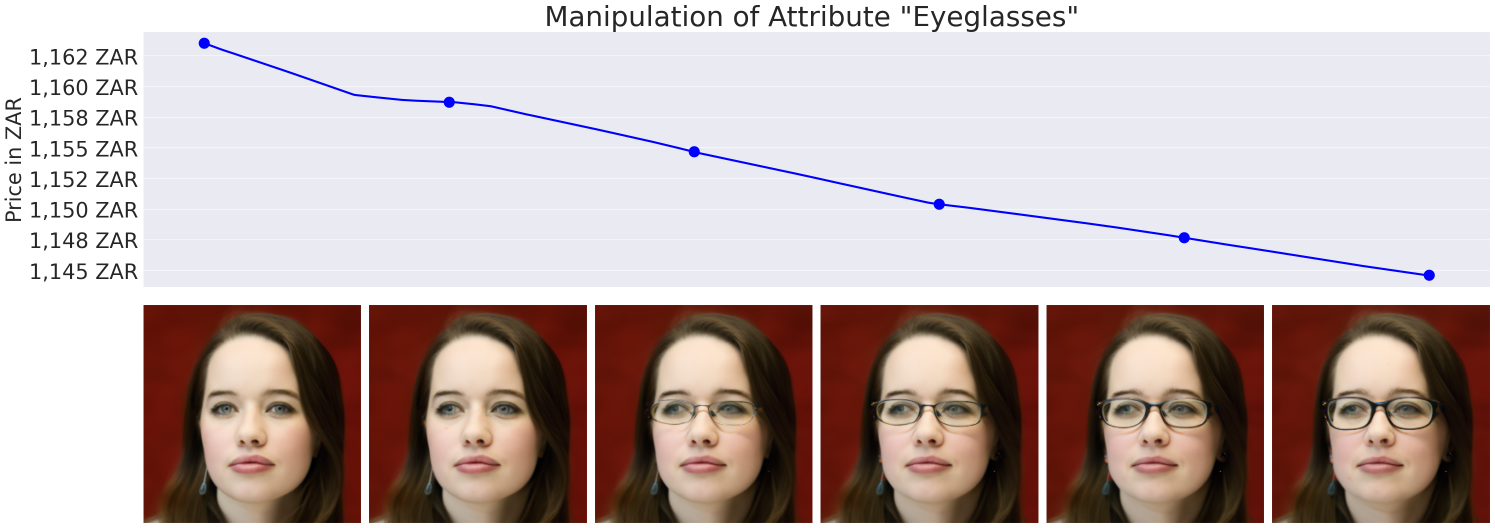}
    \label{fig:interpolation 2 }
    \vspace{0.5cm}
    \vspace{0.5cm}
  \end{subfigure}

   \caption{Effect of various facial features on the expected rental price. Through (semi) continuous interpolation, an effect trend can be visualized.}
\end{figure}
\newpage
\subsection{Descriptive Plots for the Cape Town Dataset}
\begin{figure}[!h]
    \centering
    \includegraphics[scale=0.5]{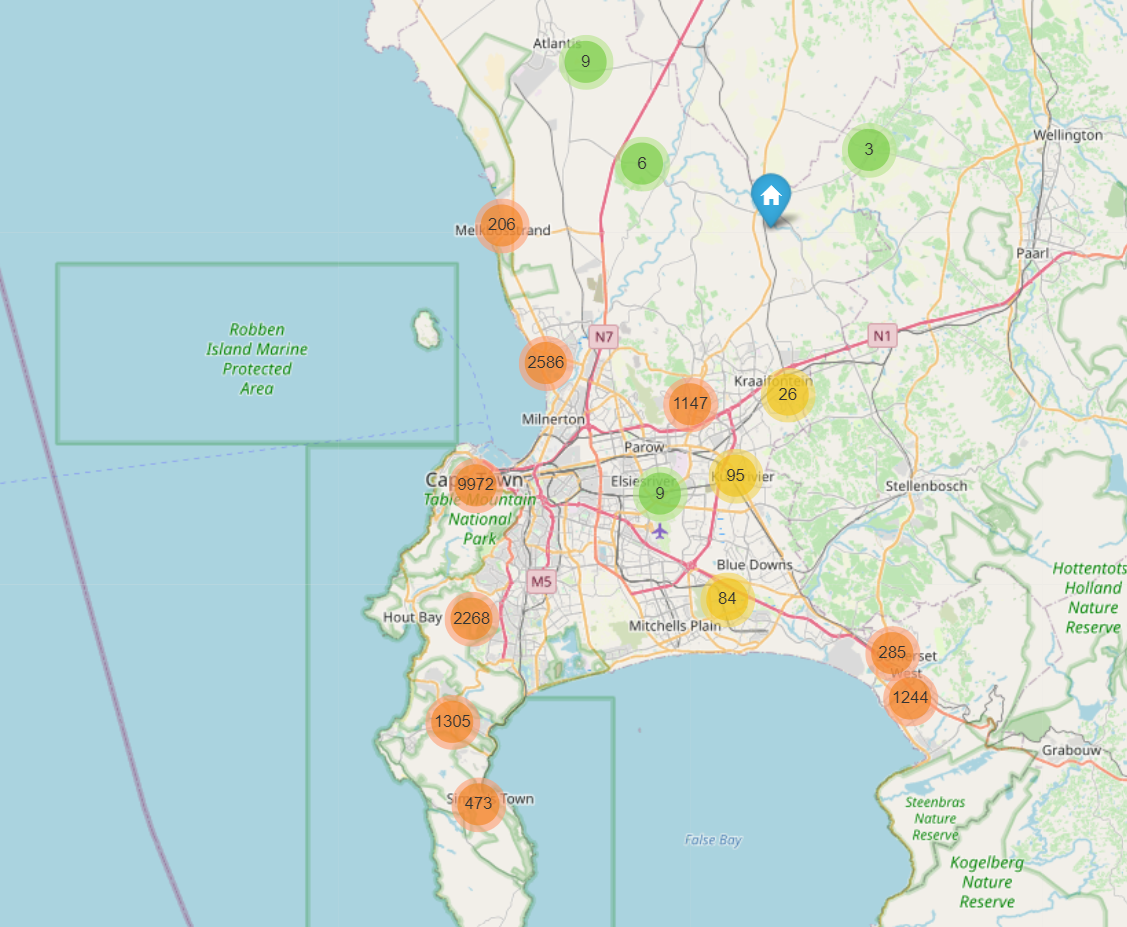}
    \caption{Caption}
    \label{fig:capetown_city}
\end{figure}

\newpage

\subsection{Explanation of Numerical Covariates for the Airbnb Datset}
\label{apx:CovariateExplanaition}
\begin{table}[H]
\begin{tabular}{|l|l|l|}
\hline
{\textbf{Covariate}} & {\textbf{Data Type}} & {\textbf{Description}} \\ \hline
{Latitude} & {Numeric} & {The latitude of the Airbnb listing's location} \\ 
{Longitude} & {Numeric} & {The longitude of the Airbnb listing's location} \\
{Room Type} & {Categorical} & {The type of room being offered by the host} \\
{Accommodates} & {Numeric} & {The maximum number of guests allowed} \\
{Bedrooms} & {Numeric} & {The number of bedrooms available} \\
{Minimum Nights} & {Numeric} & {The minimum number of nights to book} \\
{Number of Reviews} & {Numeric} & {The total number of reviews} \\
{Review Scores Value} & {Numeric} & {The average rating given to the Airbnb listing}\\
\hline
\end{tabular}
\end{table}

\newpage

\subsection{Effects of Continuous Covariates}

\begin{figure}[!h]
  \centering
  \subfloat[Global effect of the number of accommodated guests on the expected rental price for Cape Town.]{\includegraphics[width=0.48\textwidth]{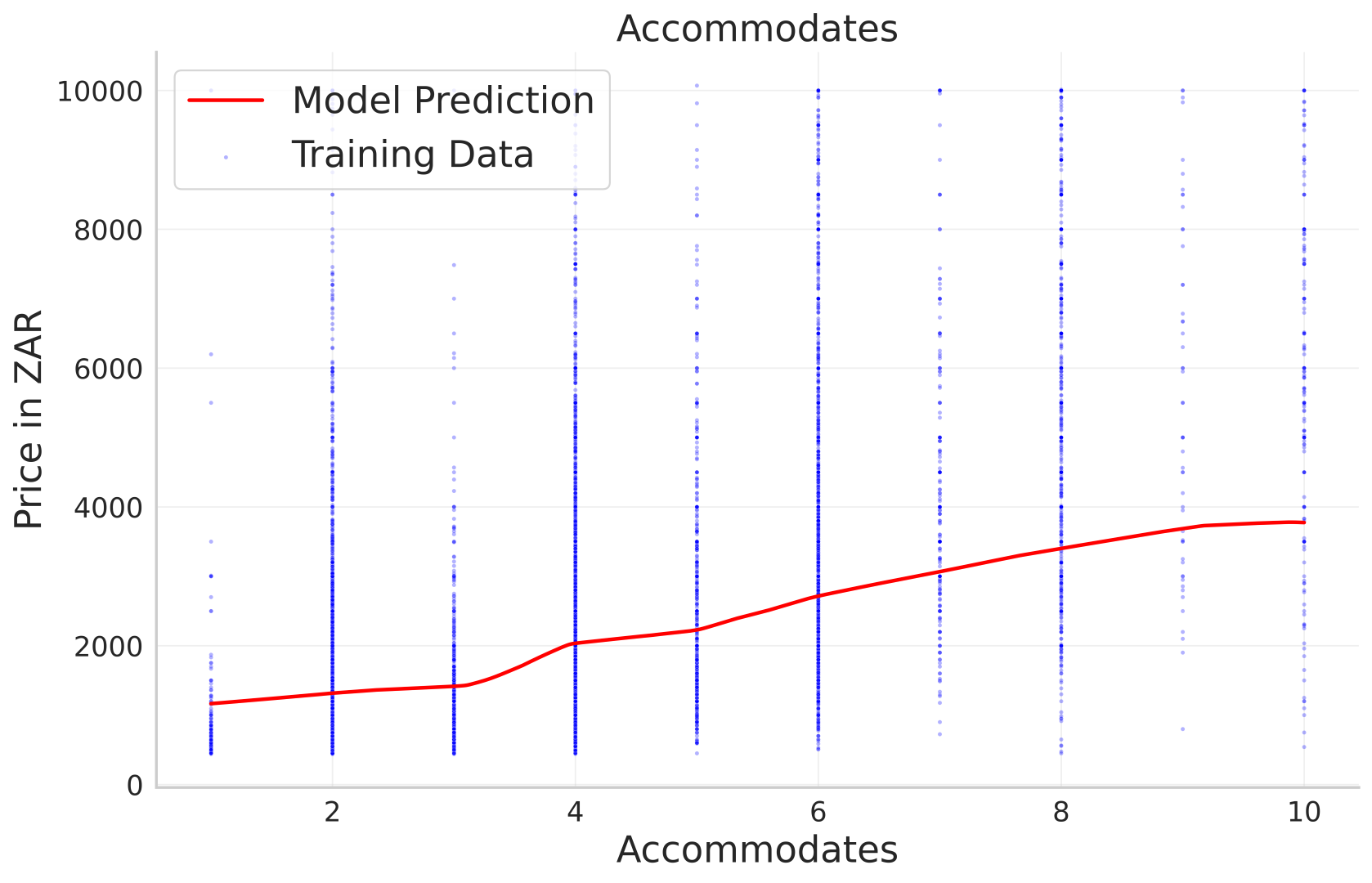}\label{fig:accomodates}}
  \hfill
  \subfloat[Global effect of number of bedrooms on the expected expected rental price for Cape Town.]
  {\includegraphics[width=0.48\textwidth]{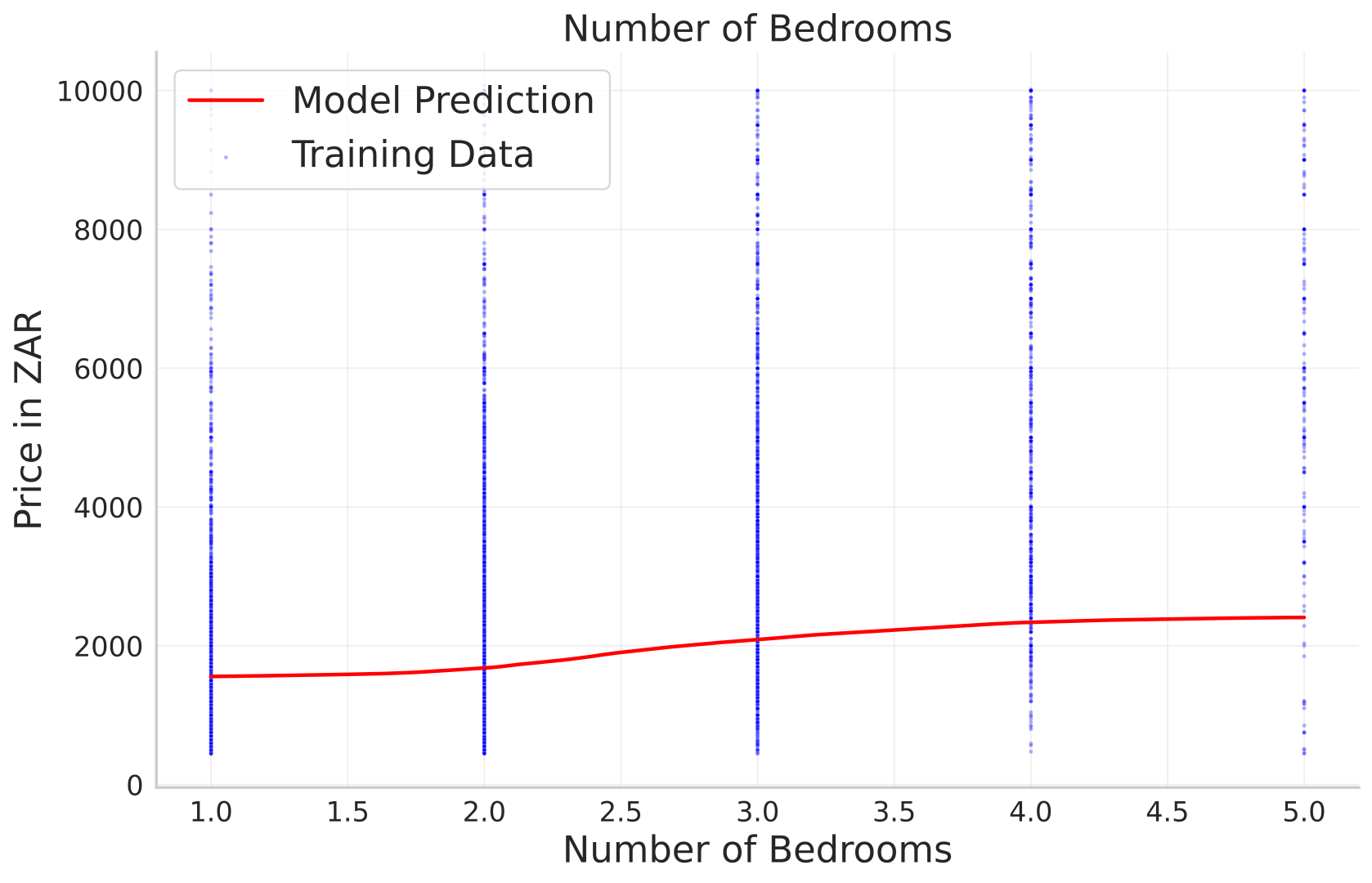}\label{fig:bedrooms}}
  \label{fig:namplots1}
  \subfloat[Global effect of the average value of review scores on the expected rental price for Cape Town.]{\includegraphics[width=0.48\textwidth]{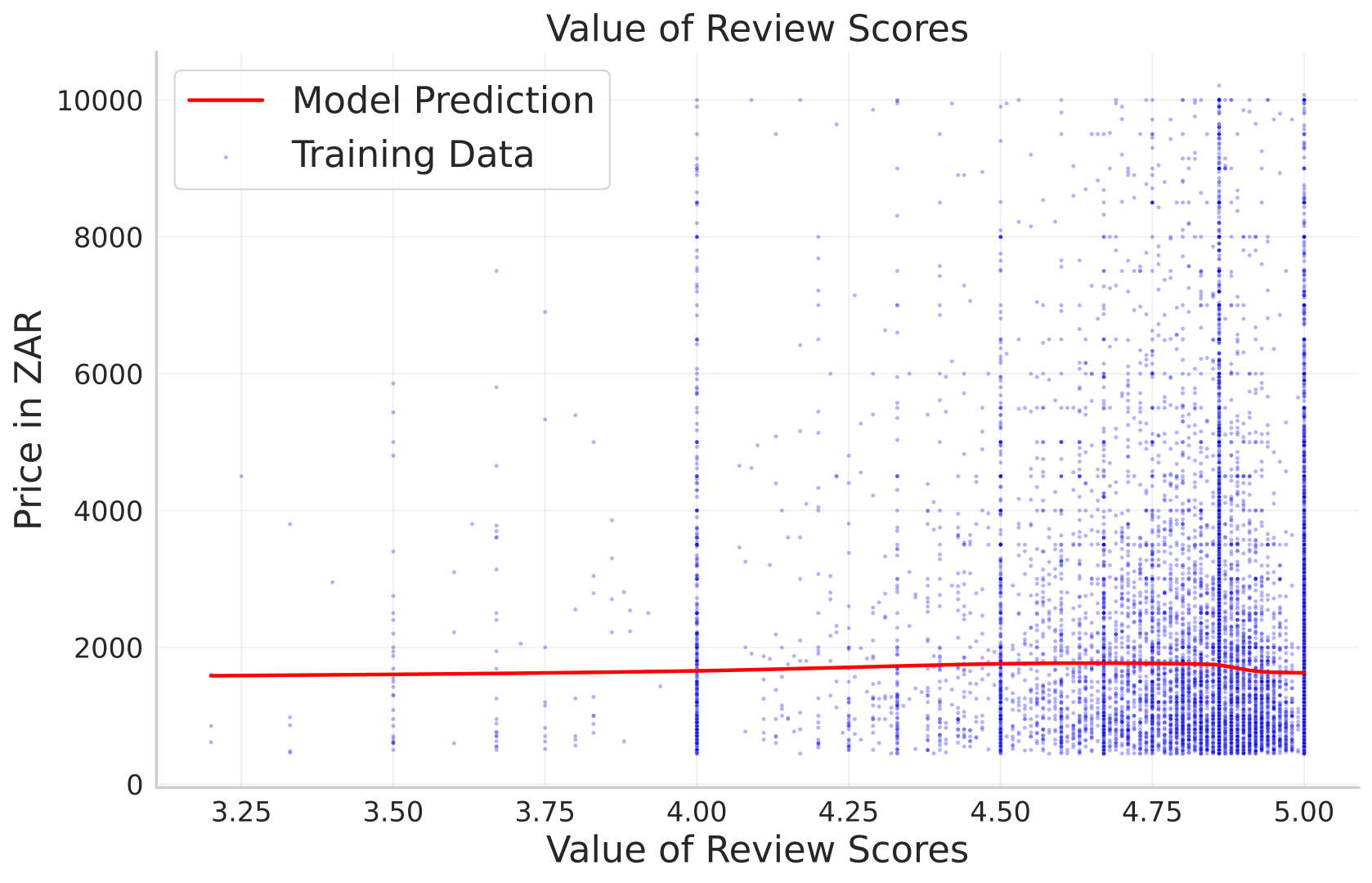}\label{fig:accomodates}}
  \hfill
  \subfloat[Global effect of the number of reviews on the expected expected rental price for Cape Town.]
  {\includegraphics[width=0.48\textwidth]{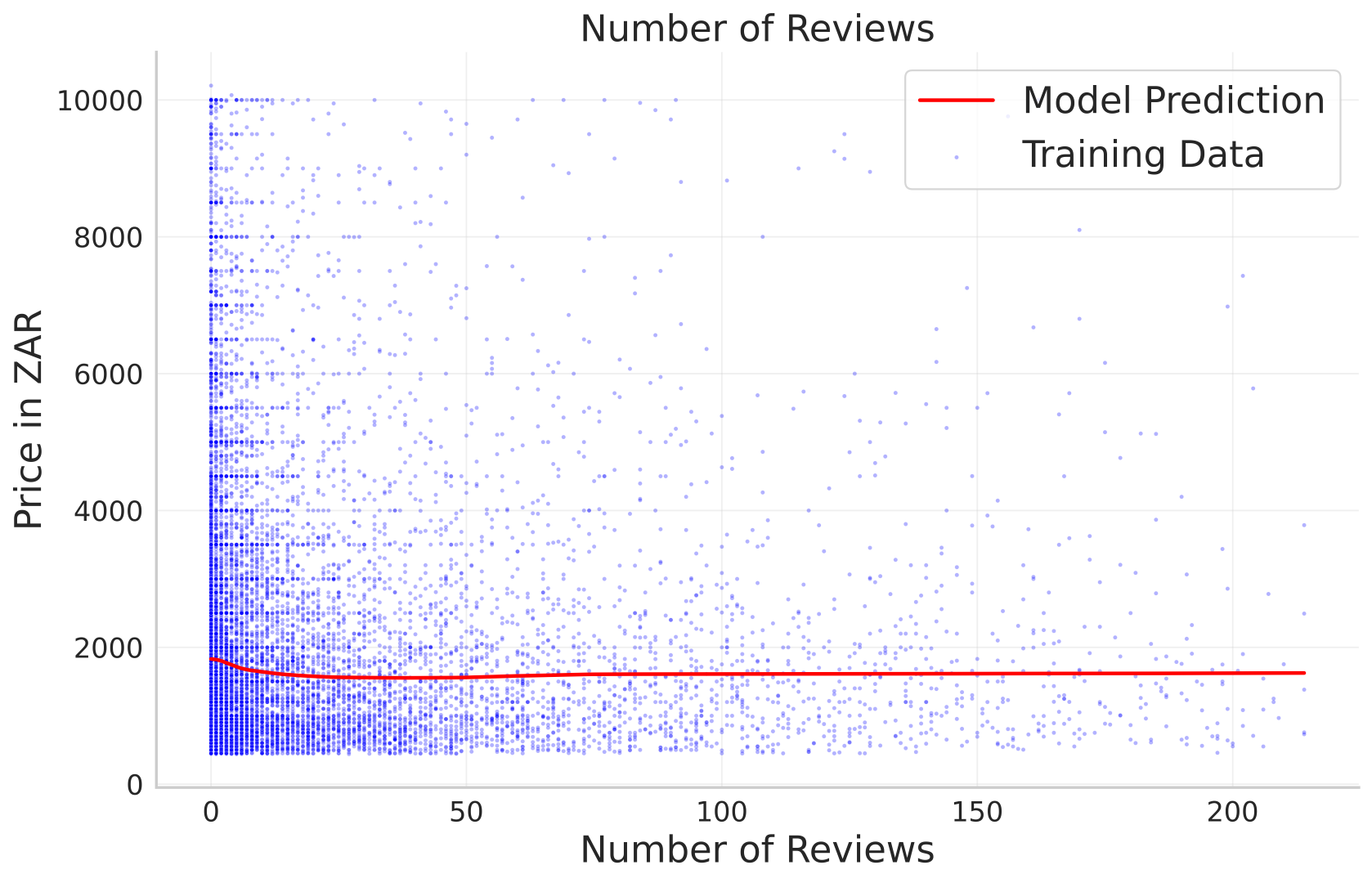}\label{fig:bedrooms}}
  \label{fig:namplots1}
  \caption{Global effects of numerical covariates}
\end{figure}

\newpage
\subsection{Global Image Effects}

\begin{figure}[!h]
  \centering
  \subfloat[Effect of image attribute "chubby".]{\includegraphics[width=0.48\textwidth]{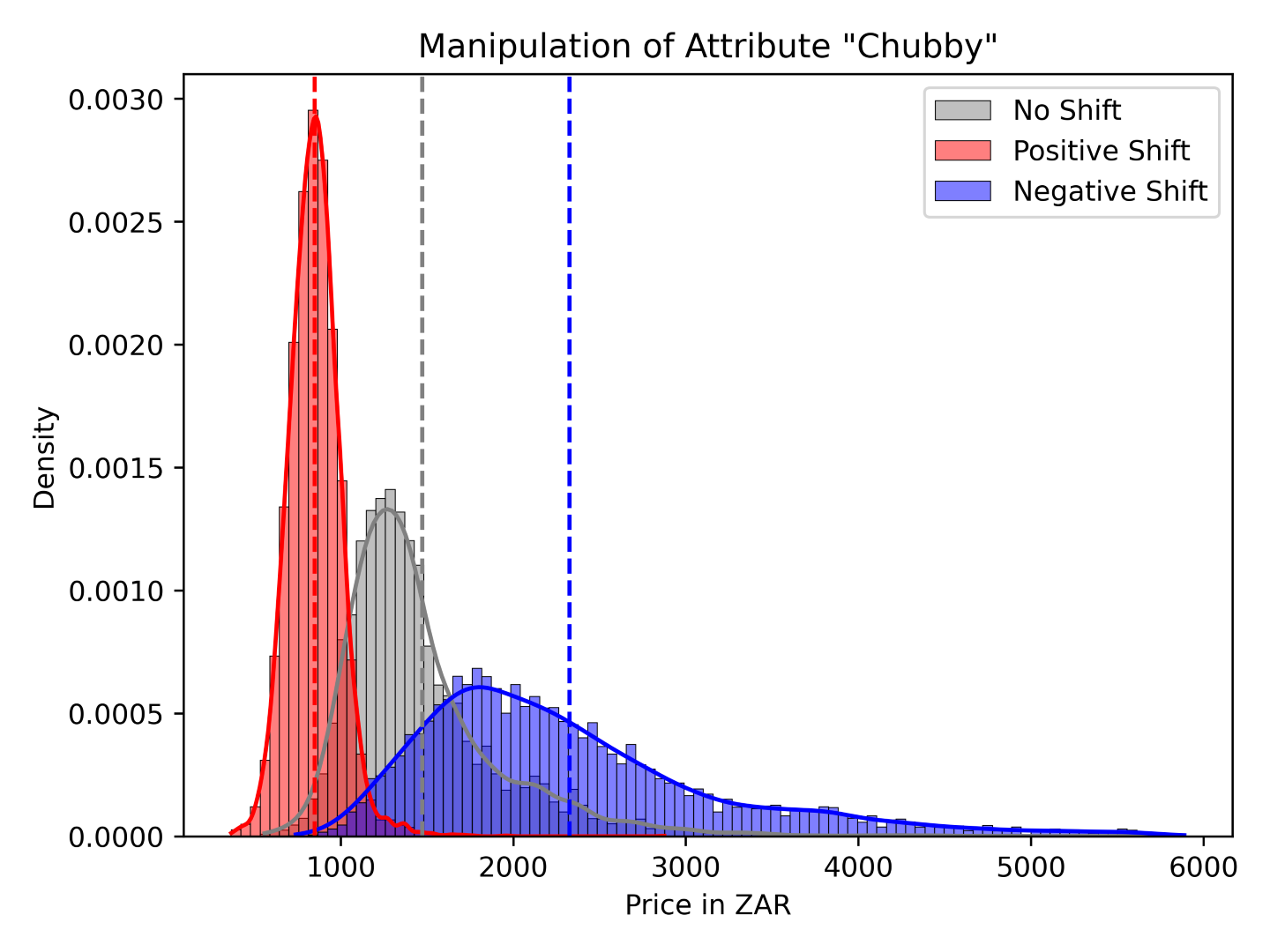}\label{fig:accomodates}}
  \hfill
  \subfloat[Effect of image attribute "young".]
  {\includegraphics[width=0.48\textwidth]{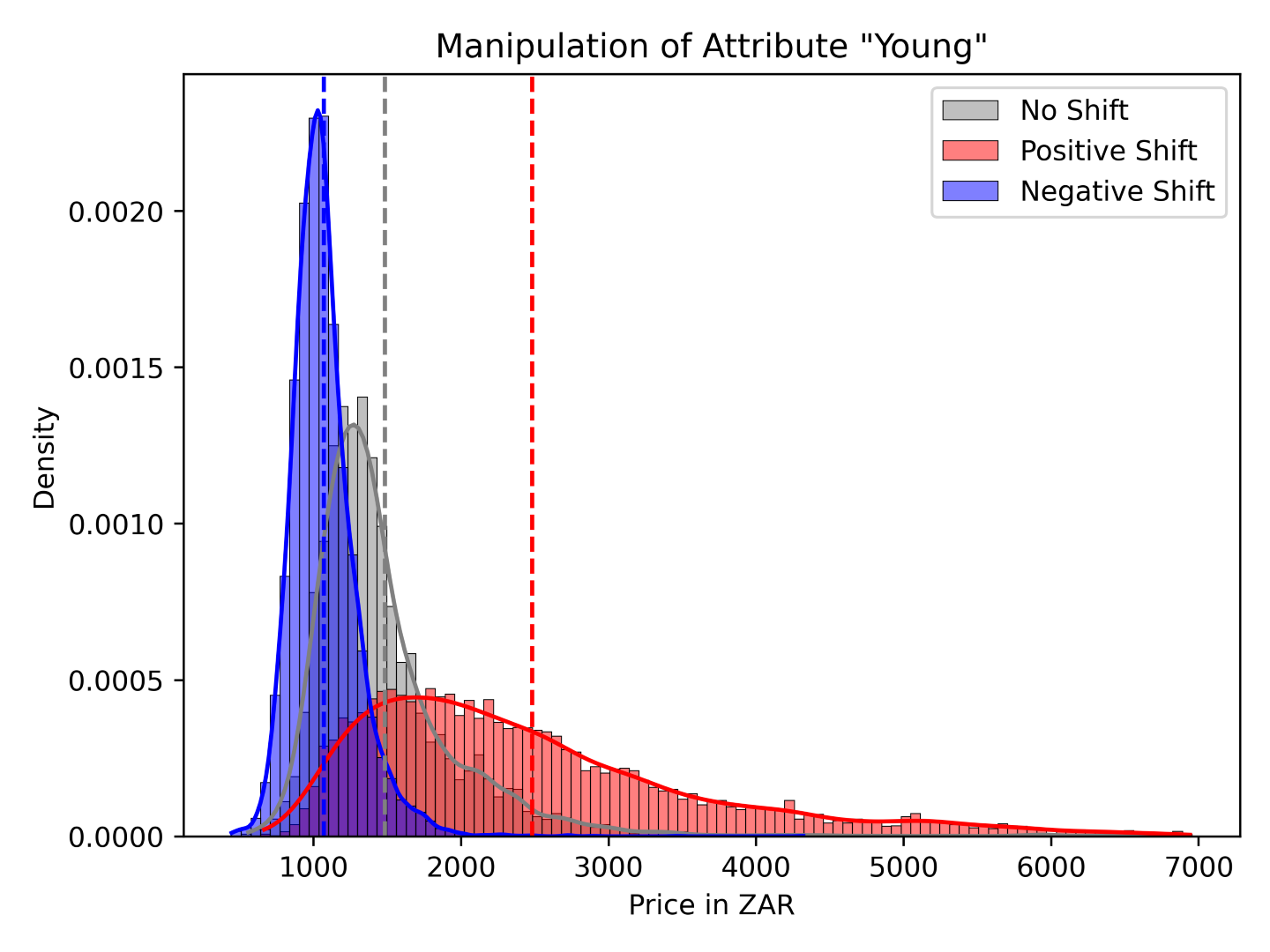}\label{fig:bedrooms}}
  \label{fig:namplots1}
  \subfloat[Effect of image attribute "smiling".]{\includegraphics[width=0.48\textwidth]{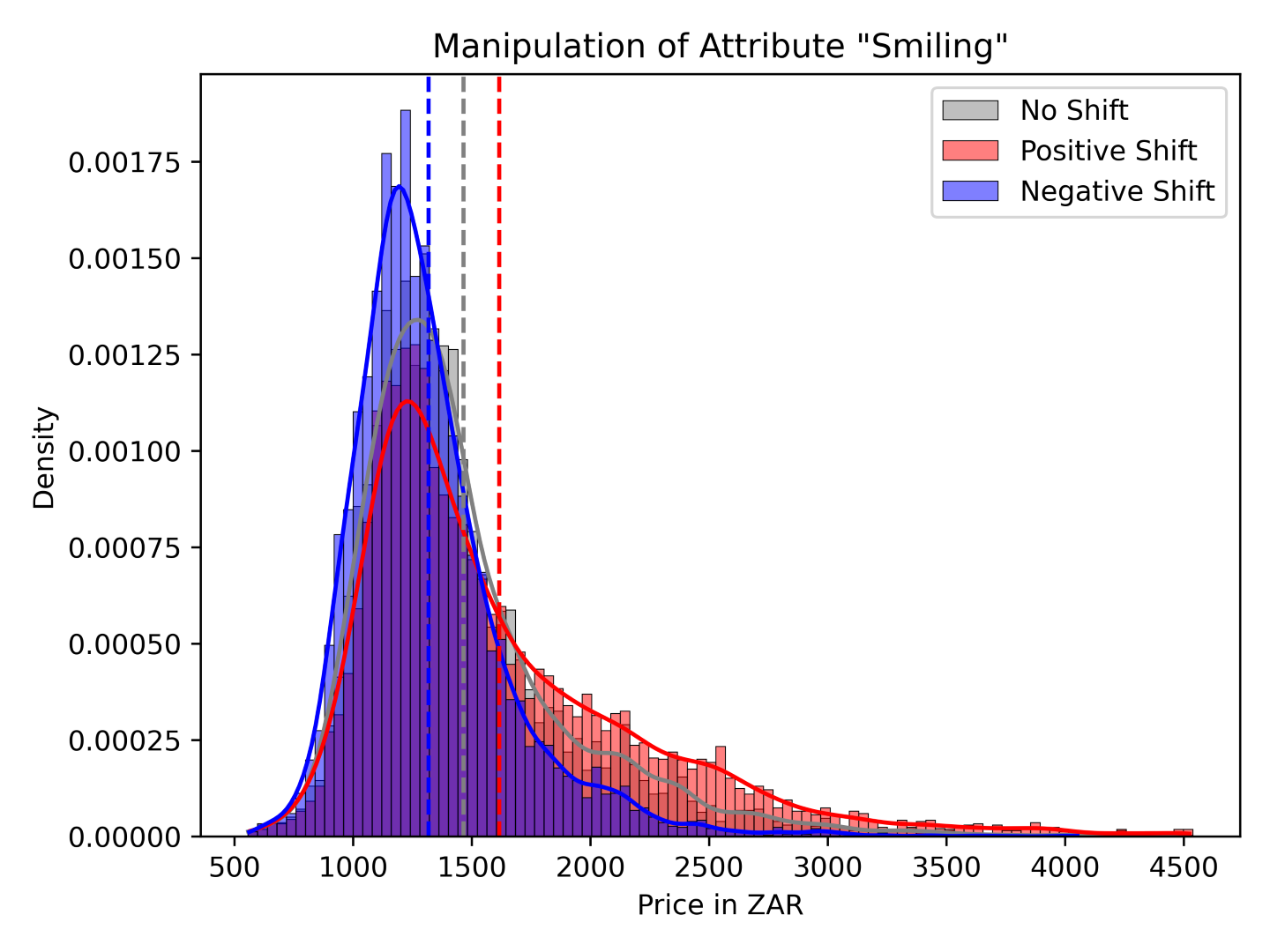}\label{fig:accomodates}}
  \hfill
  \subfloat[Effect of image attribute "pointy nose".]
  {\includegraphics[width=0.48\textwidth]{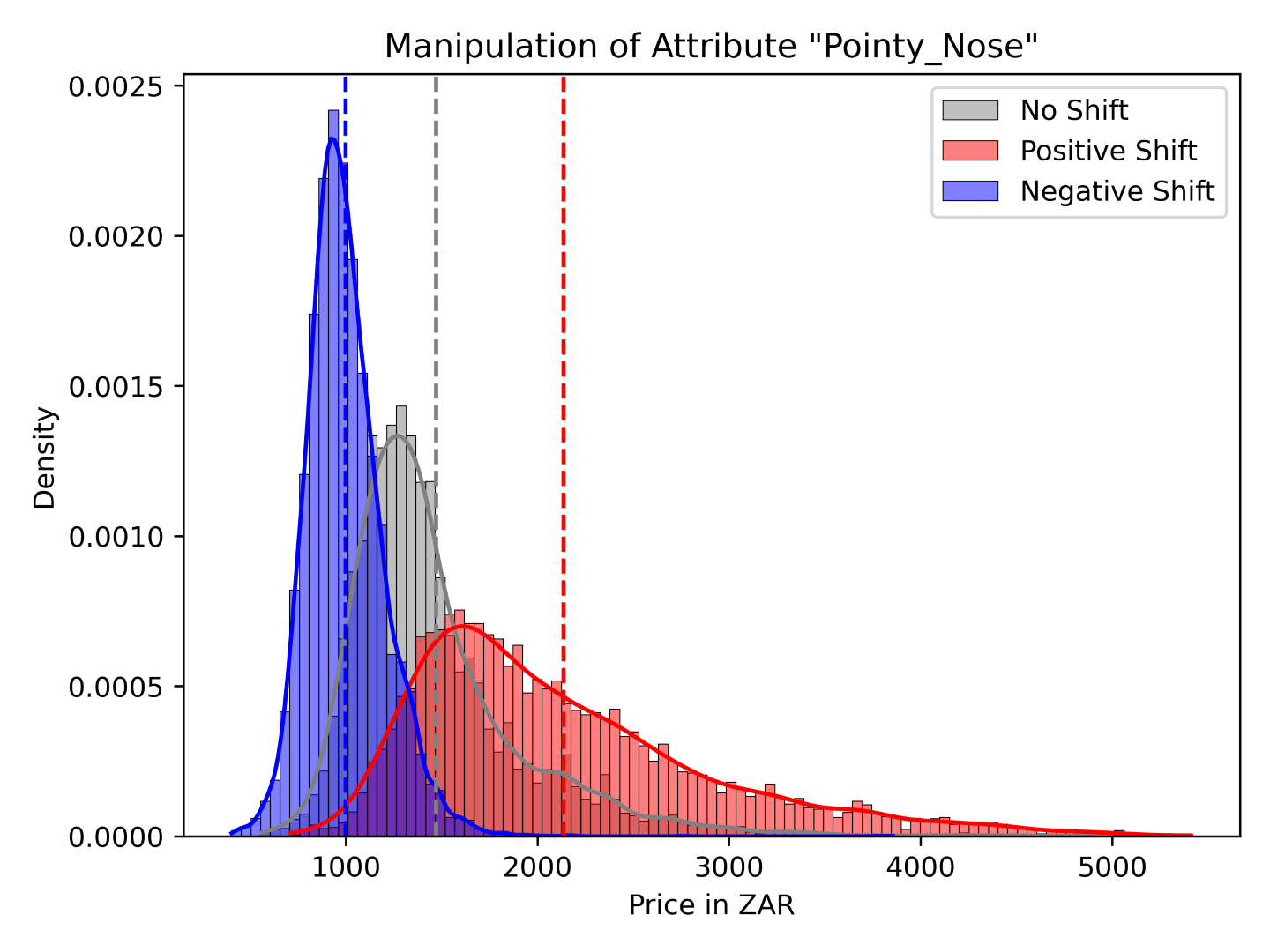}\label{fig:bedrooms}}
  \label{fig:namplots1}
  \caption{Global effects of images. The global effect of a certain image features can by analyzed through manipulating all images in the data set analogously and visualizing the shift of the predictive distribution. The grey distribution depicts the original predictive distribution without any manipulation while blue distribution depicts a negative shift and the red distribution depicts a positive shift on the semantic subcode for the chosen feature respectively. We further visualize the means of the distributions with dashed vertical lines.}
\end{figure}

\subsection{Hyperparamters} 
For the synthetic datasets, we utilize the DAE with mostly default hyperparamters \citep{preechakul2022diffusion}. We only deviate from those defaults by setting the batch size to 64 for both datasets and the learning-rate to $10^{-3}$ for the colors-data. To model the numerical effects and the effect of the embedding of the image effect by the DAE, we use the same architecture. More precisely, we utilize a four-layer MLP with skip-connections and 100 hidden units. We use dropout with a probability of 0.2 \citep{srivastava2014dropout} and feature dropout with a probability of 0.5 \citep{agarwal2021neural}. We use ReLU activation functions and optimize the model with a learning rate of $10^{-3}$ and weight decay paramter $10^{-3}$ using the Adam optimizer \citep{kingma2014adam}

For the AirBnB data, we use the DAE pretrained on Celeb-A \citep{preechakul2022diffusion}. The NAIM uses four-layer-MLPs with 100 hidden units and dropout and feature dropout set to  0.5 in the NAM component and 500 hidden units in a 4-layer MLP for the image-effect component.



\end{document}